\documentclass[conference]{IEEEtran}
\IEEEoverridecommandlockouts
\usepackage[utf8]{inputenc}
 \usepackage{algorithm}
\usepackage[noend]{algorithmic}
\usepackage{subcaption}
\usepackage{amsmath,amssymb,amsfonts}
\usepackage{algorithmic}
\usepackage{graphicx}
\usepackage{pgfplots}
\usepackage{listings}
\usepackage{bbm}
\pgfkeys{/pgf/number format/.cd,1000 sep={}}
\usepackage{tikz}
\usetikzlibrary{shapes, shadows,positioning}
\usepackage{multirow}
\usepackage{tabularx}
\usepackage{pgfplots}

\tikzstyle{element} = [draw, rounded corners, fill=gray!50]
\tikzstyle{element2} = [draw, rounded corners, fill=gray!30]
\tikzstyle{element0} = [draw, rounded corners, minimum height=2em, fill=white]
\newcommand{\set}[1]{#1}
\renewcommand{\vec}[1]{\mathbf{#1}}

\newcommand{\prob}{\mathrm{Pr}}

\newcommand{\paragraphbe}[1]{\vspace{0.03in} \noindent{\bf \em #1} }

\def\BibTeX{{\rm B\kern-.05em{\sc i\kern-.025em b}\kern-.08em
    T\kern-.1667em\lower.7ex\hbox{E}\kern-.125emX}}

\newcommand{\norm}[1]{\left\lVert#1\right\rVert}

\newlength\figureheight
\newlength\figurewidth
\newlength\trifigureheight
\newlength\trifigurewidth
\graphicspath{{./figures/}}

\renewcommand{\arraystretch}{1.1} 

\begin{document}

\title{Comprehensive Privacy Analysis of Deep Learning: Passive and Active White-box Inference Attacks against Centralized and Federated Learning}

\author{
	\IEEEauthorblockN{Milad Nasr}
	\IEEEauthorblockA{University of Massachusetts Amherst\\milad@cs.umass.edu}
	\and
\IEEEauthorblockN{Reza Shokri}
\IEEEauthorblockA{National University of Singapore\\reza@comp.nus.edu.sg}
 \and
	\IEEEauthorblockN{Amir Houmansadr}
	\IEEEauthorblockA{University of Massachusetts Amherst\\amir@cs.umass.edu}}
	
\maketitle
\thispagestyle{plain}
\pagestyle{plain}

\begin{abstract} 
Deep neural networks  are susceptible to various inference attacks as they {\em remember} information about their training data.  We design white-box inference attacks to perform a comprehensive privacy analysis of deep learning models.  We measure the privacy leakage through parameters of fully trained models as well as the parameter updates of models during training. We design inference algorithms for both centralized and federated learning, with respect to passive and active inference attackers, and assuming different adversary prior knowledge. 

We evaluate our novel {\em white-box membership inference attacks} against deep learning algorithms to trace their training data records. We show that a straightforward extension of the known black-box attacks to the white-box setting (through analyzing the outputs of activation functions) is ineffective.  We therefore design new algorithms tailored to the white-box setting by exploiting the privacy vulnerabilities of the stochastic gradient descent algorithm, which is {\em the} algorithm used to train deep neural networks.  We investigate the reasons why deep learning models may leak information about their training data.  We then show that even well-generalized models are significantly susceptible to white-box membership inference attacks, by analyzing  state-of-the-art pre-trained and publicly available models for the CIFAR dataset.  We also show how adversarial participants, in the  federated learning setting, can successfully run active membership inference attacks against other participants, even when the global model achieves high prediction accuracies.  
\end{abstract}

\section{Introduction}

Deep neural networks have shown unprecedented generalization for various learning tasks, from image and speech recognition to generating realistic-looking data.  This success has led to many applications and services that use deep learning algorithms on large-dimension (and potentially sensitive) user data, including user speeches, images, medical records, financial data, social relationships,  and location data points.  In this paper, we are interested in answering the following critical question:  What is the privacy risk of deep learning algorithms to individuals whose data is used for training deep neural networks?  In other words, how much is the information leakage of deep learning algorithms about their individual training data samples?

We define privacy-sensitive leakage of a model, about its training data, as the information that an adversary can learn from the model about them, which he is not able to infer from other models  that are trained on other data from the same distribution. This distinguishes between the information that we can learn from the model about the data population, and the information that the model leaks about the particular data samples which are in its training set.  The former indicates utility gain, and the later reflects privacy loss.  We design inference attacks to quantify such privacy leakage.

Inference attacks on machine learning algorithms fall into two fundamental and related categories: \emph{tracing} (a.k.a.\ membership inference) attacks, and \emph{reconstruction} attacks~\cite{dwork2017exposed}.  In a reconstruction attack, the attacker's objective is to infer attributes of the records in the training set~\cite{dinur2003revealing, wang2009learning}.  In a membership inference attack, however, the attacker's objective is to infer if a particular individual data record was included in the training dataset~\cite{homer2008resolving, dwork2015robust, shokri2017membership}.  This is a decisional problem, and its accuracy directly demonstrates the leakage of the model about its training data.  We thus choose this attack as the basis for our privacy analysis of deep learning models. 

Recent works have studied membership inference attacks against machine learning models in the \emph{black-box} setting, where the attacker can only observe the model predictions~\cite{shokri2017membership, yeom2018privacy}.  The results of these works show that the distribution of the training data as well as the generalizability of the model significantly contribute to the membership leakage.  Particularly, they show that overfitted models are more susceptible to membership inference attacks than generalized models. Such black-box attacks, however, might not be effective against deep neural networks that generalize well (having a large set of parameters).  Additionally, in a variety of real-world settings, the parameters of deep learning algorithms are visible to the adversaries, e.g., in a federated learning setting where multiple data holders collaborate to train a global model by sharing their parameter updates with each other through an  aggregator.

\paragraphbe{Our contributions.}  In this paper, we present a comprehensive framework for the privacy analysis of deep neural networks, using white-box membership inference attacks.  We go beyond membership inference attacks against fully-trained models.  We take all major scenarios where deep learning is used for training and fine-tuning or updating models, with one or multiple collaborative data holders, when attacker only passively observes the model updates or actively  influences the target model in order to extract more information, and for attackers with different types of prior knowledge.  Despite differences in knowledge, observation, and actions of the adversary, their objective is the same: {\em membership inference}.

A simple extension of existing black-box membership inference attacks to the  white-box setting would be using the same attack on all of the activation functions of the model.  Our empirical evaluations show that this will not result in inference accuracy better than that of a black-box attacker.  This is because the activation functions in the model tend to generalize much faster compared to the output layer.  The early layers  of a trained model extract very simple features that are not specific to the training data.  The activation functions in the last layers extract complex and abstract features, thus should contain more information about the model's training set. However, this information is more or less the same as what the output leaks about the training data.  

We design white-box inference attacks that {\bf exploit the privacy vulnerabilities of the stochastic gradient descent (SGD) algorithm}.  Each data point in the training set influences many of the model parameters, through the SGD algorithm, to minimize its contribution to the model's training loss.  The local gradient of the loss on a target data record, with respect to a given parameter, indicates how much and in which direction the parameter needs to be changed to fit the model to the data record.  To minimize the expected loss of the model, the SGD algorithm repeatedly updates model parameters in a direction that the gradient of the loss over the whole training dataset leans to {\em zero}.  Therefore, {\bf each training data sample will leave a distinguishable footprint on the gradients of the loss function over the model's parameters}.

We use the gradient vector of the model, over all parameters, on the target data point, as the main feature for the attack.  We design deep learning attack models with an architecture that processes extracted (gradient) features from different layers of the target model separately, and combines their information to compute the membership probability of a target data point.  We train the attack model for attackers with different types of background knowledge.  Assuming a subset of the training set is known to the attacker, we can train the attack model in a supervised manner. However, for the adversary that lacks this knowledge, we train the attack model in an {\bf unsupervised} manner.  We train auto-encoders to compute a membership information embedding for any data.  We then use a clustering algorithm, on the target dataset, to separate members from non-members based on their membership embedding.

To show the effectiveness of our white-box inference attack, {\bf we evaluate the privacy of pre-trained and publicly available state-of-the-art models} on the CIFAR100 dataset.  We had no influence on training these models.  Our results show that the DenseNet model\textemdash which is the best model on CIFAR100 with $82\%$ test accuracy\textemdash is not much vulnerable to black-box attacks (with a $54.5\%$ inference attack accuracy, where $50\%$ is the baseline for random guess).  However, our white-box membership inference attack obtains a considerably higher accuracy of $74.3\%$.  This shows that {\bf even well-generalized deep models might leak significant amount of information about their training data, and could be vulnerable to white-box membership inference attacks}.  

In federated learning, we show that a curious parameter server or even a participant can perform alarmingly accurate membership inference attacks against other participants.  For the DenseNet model on CIFAR100, a local participant can achieve a membership inference accuracy of $72.2\%$, even though it only observes  aggregate updates through the parameter server.  Also, the curious central parameter server can achieve a $79.2\%$ inference accuracy, as it receives the individual parameter updates from all participants. In federated learning, the {\em repeated} parameter updates of the models over different epochs on the {\em same} underlying training set is a key factor in boosting the inference attack accuracy.

As the contributions (i.e., parameter updates) of an adversarial participant can influence the parameters other parties, {\bf in the federated learning setting, the adversary can actively push SGD to leak even more information about the participants' data}.  We design an active attack that performs gradient {\em ascent} on a set of target data points before uploading and updating the global parameters.  This magnifies the presence of data points in others' training sets, in the way SGD reacts by abruptly reducing the gradient on the target data points if they are members. On the Densenet model, this leads to a $76.7\%$ inference accuracy for an adversarial participant, and a significant $82.1\%$ accuracy for an active inference attack by the central server. By isolating a participant during parameter updates, the central attacker can boost his accuracy to $87.3\%$.

\section{Inference Attacks}

\label{sec:inference_attacks}

We use membership inference attacks to measure the information leakage through deep learning models about their training data.  There are many different scenarios in which data is used for training models, and there are many different ways the attacker can observe the deep learning process.  In Table~\ref{tab:framework}, we cover the major criteria to categorize the attacks.  This includes  attack observations, assumptions about the adversary knowledge, the target training algorithm, and the mode of the attack based on the adversary's actions.  In this section, we discuss different attack scenarios as well as the techniques we use to exploit deep learning algorithms.  We also describe the architecture of our attack model, and how the adversary computes the membership probability.

\renewcommand{\arraystretch}{1.5}

\begin{table*}
    \centering
    \begin{tabular}{|l|l|p{0.76\textwidth}|}
        \hline 
        {\bf Criteria} & {\bf Attacks} & {\bf Description} \\
		\hline \hline 
        \multirow{2}{*}{Observation} 
	        & Black-box & The attacker can obtain the prediction vector $f(\vec{x})$ on arbitrary input $\vec{x}$, but cannot access the model parameters, nor the intermediate computations of $f(\vec{x})$.
	         \begin{center}
\begin{tikzpicture}
	\node [] (x) {$\vec{x}$};	
	\path (x)+(2, 0) node (f) [draw,fill=yellow!50, minimum width=6em] {$f$};
	\path (f)+(2, 0) node (fx) [] {$f(\vec{x})$};	
	
	\draw[thick,->]	(x) -- (f);
	\draw[thick,->]	(f) -- (fx);	
\end{tikzpicture}
\end{center}\\
		\cline{2-3}
	        & White-box & The attacker has access to the full model $f(\vec{x}; \vec{W})$, notably its architecture and parameters $\vec{W}$, and any hyper-parameter that is needed to use the model for predictions.  Thus, he can also observe the intermediate computations at hidden layers $h_i(\vec{x})$.
	         \begin{center}
	\begin{tikzpicture}
	\node [] (x) {$\vec{x}$};	
	\path (x)+(3.95, 0) node (f) [draw,fill=yellow!5, minimum width=20.5em, minimum height=2em] {};	
	\path (x)+(1.5, 0) node (w1) [draw,fill=yellow!20] {$\vec{W}_1$};
	\path (w1)+(1, 0) node (h1) [] {$h_1(\vec{x})$};
	\path (h1)+(1, 0) node (w2) [draw,fill=yellow!20] {$\vec{W}_2$};
	\path (w2)+(1, 0) node (h2) [] {$h_2(\vec{x})$};
	\path (h2)+(1, 0) node (dots) [] {$\cdots$};
	\path (dots)+(0.9, 0) node (wi) [draw,fill=yellow!20] {$\vec{W}_i$};
	\path (wi)+(1.5, 0) node (fx) [] {$f(\vec{x})$};
	
	\draw[thick,->]	(x) -- (f);
	\draw[->]	(h1) -- (w2);	
	\draw[->]	(h2) -- (dots);
	\draw[->]	(dots) -- (wi);	
	\draw[thick,->]	(f) -- (fx);	
	\end{tikzpicture}
\end{center}\\
        \hline \hline 
        \multirow{2}{*}{Target} 
	        & Stand-alone &  The attacker observes the final target model $f$, after the training is done (e.g., in a centralized manner) using dataset $\set{D}$.  He might also observe the updated model $f_{\Delta}$ after it has been updated (fine-tuned) using a new dataset $\set{D}_{\Delta}$. \begin{center}
	\begin{tikzpicture}
	\node [] (c) {x};
	\foreach \x in {1,2} {
		\ifthenelse{\x=1}{\newcommand\F{f}; \newcommand\D{\set{D}}}{\newcommand\F{f_{\Delta}}; \newcommand\D{\set{D}_{\Delta}}}
		
		\path (c)+(3*\x-3, 0) node (data\x) [draw,rotate=90,aspect=0.3,cylinder,fill=white,minimum width=1cm,minimum height=0.5cm] {\rotatebox{270}{$\D$}};
		
		\path (data\x)+(0, 1.2) node (f\x) [draw,fill=yellow!20,minimum width=1cm,minimum height=0.5cm] {$\F$};
		
		\draw[double,->]	(data\x) -- (f\x);
	}
	
	\draw[thick,->] (f1) -- (f2) node [midway, above, sloped] (ctext) {fine-tune};
	
	\end{tikzpicture}
\end{center}\\
        \cline{2-3}
	        & Federated &  The attacker could be the central aggregator, who observes individual updates over time and can control the view of the participants on the global parameters.  He could also be any of the participants who can observe the global parameter updates, and can control his parameter uploads. \begin{center}
	\begin{tikzpicture}
	\node [] (c) {x};
	\path (c)+(3, 3) node (server) [draw,fill=yellow!20,minimum width=3cm,minimum height=0.5] {{Aggregator } (global parameters $\vec{W}$)};
	\foreach \x in {1,2,4} {
		\path (c)+(3*\x-3, 0.6) node (A\x) [draw,fill=yellow!20,minimum width=1.8cm,minimum height=2cm] {};

		\ifthenelse{\x=4}{\newcommand\N{N}}{\newcommand\N{\x}}
				
		\path (c)+(3*\x-3, 0) node (data\x) [draw,rotate=90,aspect=0.3,cylinder,fill=white,minimum width=1cm,minimum height=0.5cm] {\rotatebox{270}{$\set{D}_\N$}};
		
		\path (data\x)+(0, 1.2) node (f\x) [draw,fill=white,minimum width=1cm,minimum height=0.5cm] {$f(\vec{x}; \vec{W}^{\{t\}}_\N)$};
		
		\draw[double,<->]	(data\x) -- (f\x);
		\draw[thick,<->] (A\x) to [out=90,in=-90] (server);
	}
	
	\path (c)+(6,0.6) node (dots) [] {$\cdots$};
	
	\path (dots)+(0,1.9) node (text) [] {down=$\vec{W}^{\{t\}}$};
	\path (dots)+(0,1.3) node (text) [] {up=$\vec{W}^{\{t\}}_i$};
	\end{tikzpicture}
\end{center} \\
        \hline \hline  
        \multirow{2}{*}{Mode} 
	        & Passive & The attacker can only observe the genuine computations by the training algorithm and the model. \\
        \cline{2-3}
	        & Active & The attacker could be one of the participants in the federated learning, who adversarially modifies his parameter uploads $\vec{W}^{\{t\}}_i$, or could be the central aggregator who adversarially modifies the aggregate parameters $\vec{W}^{\{t\}}$ which he sends to the target participant(s). \\
        \hline\hline
		\multirow{2}{*}{Knowledge} 
		& Supervised & The attacker has a data set $\set{D'}$, which contains a subset of the target set $\set{D}$, as well as some data points from the same underlying distribution as $\set{D}$ that are not in $\set{D}$. The attacker trains an inference model $h$ in a supervised manner, by minimizing the empirical loss function $\sum_{d \in \set{D'}} (1 - \mathbbm{1}_{d \in \set{D}}) h(d) + \mathbbm{1}_{d \in \set{D}} (1 - h(d))$, where the inference model $h$ computes the membership probability of any data point $d$ in the training set of a given target model $f$, i.e., $h(d) = \Pr(d \in \set{D}; f)$. \pgfplotsset{width=5cm,compat=1.8}
\pgfmathdeclarefunction{invgauss}{2}{%
	\pgfmathparse{sqrt(-2*ln(#1))*cos(deg(2*pi*#2))}%
}
\begin{center}
	\begin{tikzpicture}
	\begin{axis}[
	xtick=\empty, ytick=\empty, title={\footnotesize\em Data Universe}
	]
	\addplot [blue, only marks, mark=o, samples=50, mark size=1.2] ({invgauss(rnd/3,rnd-0.5)},{invgauss(rnd/3,rnd+0.5)});
	\addplot [black, only marks, mark=x, samples=50, mark size=1.2] ({invgauss(rnd,rnd/3+1)},{invgauss(rnd,rnd/3+1)});	
	\end{axis}
	
	\node [] (n1) {};
	\path (n1)+(3.5, 1.5) node (c) [] {};
	
	\path (c)+(3, 1) node (data) [draw,rotate=90,aspect=0.3,cylinder,fill=white,minimum width=2cm,minimum height=1cm] {\rotatebox{270}{$\set{D}$}};

	\path (c)+(3, -1) node (datap) [draw,rotate=90,aspect=0.3,cylinder,fill=red!20,minimum width=2cm,minimum height=1cm] {\rotatebox{270}{$\set{D'}$}};

	\path (c)+(3.6, 1) node (datap2) [draw,rotate=90,fill=red!20,circle] {\rotatebox{270}{$\set{D'}$}};
	
	\draw [thick, ->] (c)+(0,0.5) -- (data) node [midway, above, sloped] (ctext) {$\sim \Pr({\vec{X} = \vec{x}})$};

	\draw [thick, ->] (c)+(0,-0.5) -- (datap) node [midway, above, sloped] (ctext) {$\sim \Pr({\vec{X} = \vec{x}})$};
	    
	\end{tikzpicture}
\end{center}\\
		\cline{2-3}
		& Unsupervised &  The attacker has data points that are sampled from the same underlying distribution as $\set{D}$.  However, he does not have information about whether a data sample has been in the target set $\set{D}$. \\
		\hline
    \end{tabular}
    \caption{\small Various categories of inference attacks against machine learning models, based on their prior knowledge, observation, mode of attack, and the training architecture of the target models.}
    \label{tab:framework}
\end{table*}
\begin{figure}[t!]
	\centering
\begin{tikzpicture}[scale=0.9, every node/.style={transform shape}]
	\node [] (x) {$\vec{x}$};	
	\path (x)+(3.83, 0) node (model) [element0, dotted, fill=yellow!20, minimum width=20.5em, minimum height=9em] {};

	\path (model)+(0, -2) node (modellabel) [] {{\em target model}};
		
	\path (x)+(0.4, 0) node (input) [element2, rotate=90, minimum width=6em] {};
	\path (input)+(0.85, 0) node (param1) [] {$\vec{W}_{1}$};
	\path (param1)+(0.85, 0) node (hidden1) [element2, rotate=90, minimum width=8em] {};
	\path (hidden1)+(0.85, 0) node (param2) [] {$\vec{W}_{2}$};
	\path (param2)+(0.85, 0) node (hidden2) [element2, rotate=90, minimum width=7em] {};
	\path (hidden2)+(1.35, 0) node (dots) [] {$\cdots$};
	\path (dots)+(1.25, 0) node (parami) [] {$\vec{W}_{i}$};
	\path (parami)+(0.85, 0) node (outputlayer) [element2, rotate=90, minimum width=4em] {};
	\path (outputlayer)+(0.83, 0) node (lossfxy) [rotate=90] {$L(f(\vec{x}; \vec{W})), y$};	
	\path (lossfxy)+(0.83, 0) node (label) [rotate=90] {$\vec{y}$};

	\path (param1)+(-0.1, 3.1) node (grad1g) [gray,element0] {$\frac{\partial L}{\partial \vec{W}_{1}}$};
	
	\path (param1)+(0, 3) node (grad1) [element0] {$\frac{\partial L}{\partial \vec{W}_{1}}$};

	\path (param2)+(-0.1, 3.1) node (grad2g) [gray,element0] {$\frac{\partial L}{\partial \vec{W}_{2}}$};
	
	\path (param2)+(0, 3) node (grad2) [element0] {$\frac{\partial L}{\partial \vec{W}_{2}}$};
	
	\path (dots)+(-0.3, 3) node (dotsi) [] {$\cdots$};	

	\path (parami)+(-0.1, 3.1) node (gradig) [gray,element0] {$\frac{\partial L}{\partial \vec{W}_{i}}$};
	
	\path (parami)+(0, 3) node (gradi) [element0] {$\frac{\partial L}{\partial \vec{W}_{i}}$};
	
	\path (lossfxy)+(-0.1, 3.1) node (lossg) [gray,element0] {$L$};
		
	\path (lossfxy)+(0, 3) node (loss) [element0] {$L$};
		
	\path (hidden1)+(-0.1, 4.1) node (output1g) [gray,element0] {$h_1(\vec{x})$};

	\path (hidden1)+(0, 4) node (output1) [element0] {$h_1(\vec{x})$};
	
	\path (hidden2)+(-0.1, 4.1) node (output2g) [gray,element0] {$h_2(\vec{x})$};
	
	\path (hidden2)+(0, 4) node (output2) [element0] {$h_2(\vec{x})$};

	\path (dots)+(0.3, 4) node (dotsi) [] {$\cdots$};	

	\path (outputlayer)+(-0.1, 4.1) node (outputg) [gray,element0] {$f(\vec{x})$};
	
	\path (outputlayer)+(0, 4) node (output) [element0] {$f(\vec{x})$};
	
	\draw[double,->,gray]	(param1) -- (grad1);
	\draw[double,->,gray]	(param2) -- (grad2);
	\draw[double,->,gray]	(parami) -- (gradi);
	\draw[double,->,gray]	(lossfxy) -- (loss);
	\draw[double,->,gray]	(hidden1) -- (output1);
	\draw[double,->,gray]	(hidden2) -- (output2);
	\draw[double,->,gray]	(outputlayer) -- (output);

	\path (grad1)+(3.85, 5.7) node (attack) [element0, dotted, fill=red!20, minimum width=25em, minimum height=22.5em] {};
	
	\path (attack)+(-5, 0) node (attacklabel) [rotate=90] {{\em attack model}};

	\path (attacklabel)+(0, -5.5) node (attacklabel) [rotate=90] {{\em attack features}};

	\path (param1)+(0, 5.5) node (cp1) [element, minimum width=2em, minimum height=3.5em] {{\footnotesize CNN}};
	\path (hidden1)+(0, 5.5) node (ch1) [element, minimum width=2em, minimum height=3.5em] {{\footnotesize FCN}};
	\path (param2)+(0, 5.5) node (cp2) [element, minimum width=2em, minimum height=3.5em] {{\footnotesize CNN}};
	\path (hidden2)+(0, 5.5) node (ch2) [element, minimum width=2em, minimum height=3.5em] {{\footnotesize FCN}};
	\path (parami)+(0, 5.5) node (cpi) [element, minimum width=2em, minimum height=3.5em] {{\footnotesize CNN}};
	\path (dots)+(0, 5.5) node (di) [] {$\cdots$};
	\path (outputlayer)+(0, 5.5) node (co) [element, minimum width=2em, minimum height=3.5em] {{\footnotesize FCN}};	
	\path (lossfxy)+(0, 5.5) node (cl) [element, minimum width=2em, minimum height=3.5em] {{\footnotesize FCN}};	
	\path (label)+(0, 5.5) node (cy) [element, minimum width=2em, minimum height=3.5em] {{\footnotesize FCN}};	

	\draw[thick,->]	(output1) -- (ch1);
	\draw[thick,->]	(output2) -- (ch2);
	\draw[thick,->]	(output) -- (co);
	\draw[thick,->]	(grad1) -- (cp1);
	\draw[thick,->]	(grad2) -- (cp2);
	\draw[thick,->]	(gradi) -- (cpi);
	\draw[thick,->]	(loss) -- (cl);
	\draw[thick,->]	(label) -- (cy);

	\path (cp1)+(3.85, 2) node (call) [element, minimum width=24em, minimum height=4em] {Encoder {\footnotesize (FCN)}};
	
	\path (cp1)+(0, 0.6) node (cp1d) [] {};
	\path (cp2)+(0, 0.6) node (cp2d) [] {};
	\path (cpi)+(0, 0.6) node (cpid) [] {};
	\path (ch1)+(0, 0.6) node (ch1d) [] {};
	\path (ch2)+(0, 0.6) node (ch2d) [] {};
	\path (co)+(0, 0.6) node (cod) [] {};
	\path (cl)+(0, 0.6) node (cld) [] {};

	\path (call)+(0, 1.2) node (out) [element] {};	
	
	\path (out)+(-1.3,0) node (outlabel) [] {{\bf\em attack output}:};

	\path (call)+(0, 3.4) node (unsupervised) [element0, dotted, fill=red!10, minimum width=24em, minimum height=9.5em] {};

	\path (out)+(0, 1.2) node (decoder) [element, minimum width=23em, minimum height=3em] {Decoder {\footnotesize (FCN)}};

	\path (decoder)+(0, 2.4) node (unsupervisedlabel) [] {{\em unsupervised attack component}};
	
	\path (decoder)+(-3.25,1.5) node (dout1) [element0] {$L$};
	\path (dout1)+(1.5,0) node (dout2) [element0] {$\mathbbm{1}_{y = \arg\max f(\vec{x})}$};
	\path (dout2)+(1.8,0) node (dout3) [element0] {$f(\vec{x})_{y}$};	
	\path (dout3)+(1.35,0) node (dout4) [element0] {$\mathrm{H}(f(\vec{x}))$};
	\path (dout4)+(1.4,0) node (dout5) [element0] {$\norm{\frac{\partial L}{\partial \vec{W}}}$};

	\path (decoder)+(0,1.1) node (decoderout) [] {};
	\draw[double,->]	(decoder) -- (decoderout);

\end{tikzpicture}
	\caption{\small The architecture of our white-box inference attack.  Given target data $(\vec{x}, y)$, the objective of the attack is to determine its membership in the training set $\set{D}$ of target model $f$.  The attacker runs the target model $f$ on the target input $\vec{x}$, and computes all the hidden layers $h_i(\vec{x})$, the model's output $f(\vec{x})$, and the loss function $L(f(\vec{x}), y; \vec{W})$, in a forward pass.  The attacker also computes the gradient of the loss with respect to the parameters of each layer $\frac{\partial L}{\partial \vec{W}_i}$, in a backward pass.  These computations, in addition to the one-hot encoding of the true label $\vec{y}$, construct the input features of the inference attack.  The attack model consists of convolutional neural network (CNN) components and fully connected network (FCN) components.  For attacking federated learning and fine-tuning, the attacker observes each attack feature $T$ times, and stacks them before they are passed to the corresponding attack component. For example, the loss features are composed as $L = \{L^{\{1\}}, L^{\{2\}}, \cdots, L^{\{T\}}\}$). The outputs of the CNN and FCN components are appended together, and this vector is passed to a fully connected encoder.  The output of the encoder, which is a single value, is the attack output.  This represents an embedding of the membership information in a single value.  In the supervised attack setting, this embedding is trained to be $\prob\{(\vec{x}, y) \in \set{D}\}$.  In the unsupervised setting, a decoder is trained to reconstruct important features of the attack input (such as the model's output uncertainty $\mathrm{H}(f(\vec{x}))$, and the norm of its gradients $\norm{\frac{\partial L}{\partial \vec{W}}}$) from the attack output.  This is similar to deep auto-encoders.  All unspecified attack layers are fully connected. The details of the architecture of the attack is presented in Table~\ref{tab:model_sizes} in Appendix~\ref{sec:appendix}.\\[-15pt] }
	\label{fig:attack_model}
\end{figure}
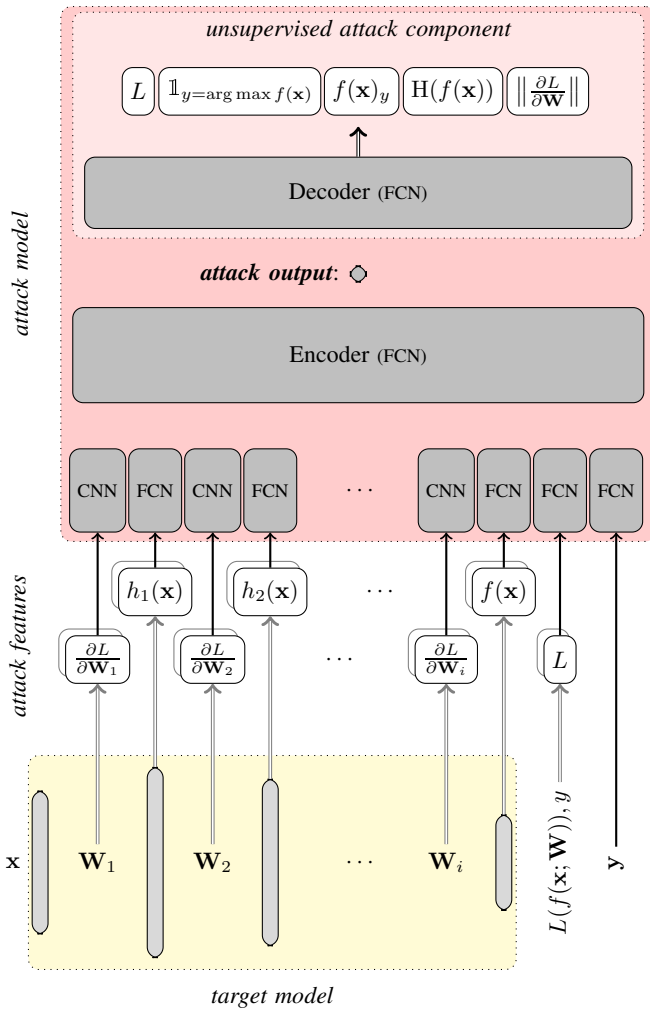

\subsection{Attack Observations: Black-box vs.\ White-box Inference}

\label{sec:inference_blackbox_whitebox}

The adversary's observations of the deep learning algorithm are what constitute the inputs for the inference attack.

\paragraphbe{Black-box.}  In this setting, the adversary's observation is limited to the output of the model on arbitrary inputs.  For any data point $\vec{x}$, the attacker can only obtain $f(\vec{x}; \vec{W})$.  The parameters of the model $\vec{W}$ and the intermediate steps of the computation are \emph{not} accessible to the attacker.  This is the setting of machine learning as a service platforms.  Membership inference attacks against black-box models are already designed, which exploit the statistical differences between a model's predictions on its training set versus unseen data~\cite{shokri2017membership}.

\paragraphbe{White-box.}  In this setting, the attacker obtains the model $f(\vec{x}; \vec{W})$ including its parameters which are needed for prediction.  Thus, for any input $\vec{x}$, in addition to its output, the attacker can compute all the intermediate computations of the model.  That is, the adversary can compute any function over $\vec{W}$ and $\vec{x}$ given the model.  The most straightforward functions are the outputs of the hidden layers, $h_i(\vec{x})$ on the input $\vec{x}$.  As a simple extension, the attacker can extend black-box membership inference attacks (which are limited to the model's output) to the outputs of all activation functions of the model.  However, this does not necessarily contain all the useful information for membership inference.  Notably, the model output and activation functions could generalize if the model is well regularized.  Thus, there might not be much difference, in distribution, between the activation functions of a model on its training versus unseen data.  This can significantly limit the power of the inference attacks (as we also show in our evaluations).

What we suggest is to exploit the  algorithm  used to train  deep learning models: the stochastic gradient descent (SGD) algorithm.  Let $L(f(\vec{x}; \vec{W}), y)$ be the loss function for the classification model $f$.  During the training, the SGD algorithm minimizes the empirical expectation of the loss function over the training set $\set{D}$:
\begin{align}
\min_{\vec{W}} \mathbb{E}_{(\vec{x},y) \sim {\set{D}}} \big[L(f(\vec{x}; \vec{W}), y) \big]
\end{align}

The SGD algorithm solves this minimization by repeatedly updating parameters, $\vec{W}$, towards reducing the loss on small randomly selected subsets of $\set{D}$.  Thus, for any data record in the training dataset, the gradient of the loss $\frac{\partial L}{\partial \vec{W}}$ over the data record is pushed towards zero, after each round of training.  This is exactly what we can exploit to extract information about a model's training data.  

For a target data record $(\vec{x}, y)$, the adversary can compute the loss of the model $L(f(\vec{x}; \vec{W}), y)$, and  can compute the gradients of the loss with respect to all parameters $\frac{\partial L}{\partial \vec{W}}$ using a simple back-propagation algorithm.  Given the large number of parameters used in deep neural networks (millions of parameters),  the vector with such a significantly large dimension cannot properly generalize over the training data (which in many cases is an order of magnitude smaller in size).  Therefore, the distribution of the model's gradients on members of its training data, versus non-members, is likely to be distinguishable.  This can help the adversary to run an accurate membership inference attack, even though the classification model (with respect to its predictions) is well-generalized.

\paragraphbe{Inference model.}  We illustrate the membership inference attack in Figure~\ref{fig:attack_model}.  The significance of gradient (as well as activation) computations for a membership inference attack varies over the layers of a deep neural network.  The first layers tend to contain less information about the specific data points in the training set, compared to non-member data record from the same underlying distribution.  We can provide the gradients and activations of each layer as separate inputs to the attacker, as the attacker might need to design a specific attack for each layer.  This enables the inference attack to split the inference task across different layers of the model, and then combine them to determine the membership.  This engineering of the attack model architecture empowers the inference attack, as it reduces the capacity of the attack model and helps finding the optimal attack algorithm with less background data. 

The distinct inputs to the attack model are the set of gradients $\frac{\partial L}{\partial \vec{W}_1}, \frac{\partial L}{\partial \vec{W}_2}, \cdots$, the set of activation vectors for different layers $h_1(\vec{x}), h_2(\vec{x}), \cdots$, the model output $f(\vec{x})$, the one-hot encoding of the label $\vec{y}$, and the loss of the model on the target data $L(f(\vec{x}; \vec{W}), y)$.  Each of these are separately fed into the attack model, and are analyzed separately using independent components.

\paragraphbe{Inference attack components.}  The attack model is composed of feature extraction components and an encoder component.  To extract features from the output of each layer, plus the one-hot encoding of the true label and the loss, we use fully connected network (FCN) submodules with one hidden layer.  We use convolutional neural network (CNN) submodules for the gradients.  When the gradients are computed on fully connected layers (in the target model), we set the size of the convolutional kernel to the input size of the fully connected layer, to capture the correlation of the gradients in each activation function.  We reshape the output of each submodule component into a flat vector, and then concatenate the output of all components.
We combine the outputs of all attack feature extraction components using a fully connected encoder component with multiple hidden layers.  The output of the encoder is a single score, which is the output of the attack.  This score (in the supervised attack raining) predicts the membership probability of the input data.

\subsection{Inference Target: Stand-alone vs.\ Federated Learning}

\label{sec:inference_standalone_federated}

There are two major types of training algorithms for deep learning, depending on whether the training data is available all in one place (i.e., stand-alone centralized training), or it is distributed among multiple parties who do not trust each other (i.e., federated learning)~\cite{konevcny2016federated}.  In both cases, the attacker could be the entity who obtains the final trained model.  In addition to such attack setting, the attacker might observe an updated version of the model after fine-tuning, for instance, which is very common in deep learning.   Besides, in the case of federated learning, the attacker can be an entity who {\em participates} in the training.  The settings of fine-tunning and federated learning are depicted in Table~\ref{tab:framework}.

\paragraphbe{Stand-alone fine-tunning.}   A model $f$ is trained on dataset $\set{D}$.  At a later stage it is updated to $f_\Delta$ after being fine-tuned using a new dataset $\set{D}_\Delta$.  If the attacker observes the final outcome, we want to measure the information leakage of the final model $f_\Delta$ about the whole training set $\set{D}\cup\set{D}_\Delta$.  However, given that two versions of the model exist (before and after fine-tuning), we are also interested in measuring the extra information that could be learned about the training data, from the two model snapshots.  The attacker might also be interested only in recovering information about the new set $\set{D}_\Delta$.  This is very relevant in numerous cases where the original model is trained using some unlabeled (and perhaps public) data, and then it is fine-tunned using sensitive private labeled data.

The model for inference attacks against fine-tunned models is a special case of our membership inference model for attacking federated learning.  In both cases, the attacker observes multiple versions of the target model.

\paragraphbe{Federated learning.}  In this setting, $N$ participants, who have different training sets $\set{D}_i$, agree on a single deep learning task and model architecture to train a global model.  A central server keeps the latest version of the parameters $\set{W}$ for the global model.  Each participant has a local model, hence a local set of parameters $\set{W}_i$.  In each epoch of training, each participant downloads the global parameters, updates them locally using SGD algorithm on their local training data, and uploads them back to the server.  The parameter server computes the average value for each parameter using the uploaded parameters by all participants. This collaborative training continues until the global model converges. 

There are two possibilities for the position of the attacker in federated learning:  The adversary can be the centralized parameter server, or one of the participants.  A curious parameter server can receive updates from each individual participant over time $\set{W}_i^{\{t\}}$, and use them to infer information about the training set of each participant.  A malicious parameter server can also control the view of each participant on the global model, and can act actively to extract more information about the training set of a participant (as we discuss under active attacks).  Alternatively, the adversary can be one of the participants.  An adversarial participant can only observe the global parameters over time $\set{W}^{\{t\}}$, and craft his own adversarial parameter updates $\set{W}_i^{\{t\}}$ to gain more information about the union of the training data of all other participants. 

In either of these cases, the adversary observes multiple versions of the target model over time.  The adversary can try to run an independent membership inference attack on each of these models, and then combine their results.  This, however, might not capture the dependencies between parameter values over time, which can leak information about the training data.  Instead, in our design we  make use of a single inference model, where each attack component (e.g., components for gradients of layer $i$) processes all of its corresponding inputs over the observed models at once.  This is illustrated in Figure~\ref{fig:attack_model}.  For example, for the attack component that analyzes the loss value $L$, the input dimension can be $1 \times T$, if the adversary observes $T$ versions of the target model over time.  The output of the attack component is also $T$ times larger than the case of attacking a stand-alone model.  These correlated outputs, of all attack components, are processed all at once by the inference model.

\subsection{Attack Mode: Passive vs.\ Active Inference Attack}

\label{sec:inference_passive_active}

The inference attacks are mostly passive, where the adversary makes observations without modifying the learning process.  This is the case notably for attacking models after the training is over, e.g., the stand-alone setting. 

\paragraphbe{Active attacks.} The adversary, who is participating in the training process, can actively influence the target model in order to extract more information about its training set.  This could be the case notably for running inference attacks against federated learning.  In this setting, the central parameter server or a curious participant can craft adversarial parameter updates for a follow-up inference attack.  The inference model architecture will be the same for passive and active attacks.

The active attacker can exploit the SGD algorithm to run the active attack.  The insight we use to design our attack is that the SGD algorithm forcefully decreases the gradient of the loss on the training data, with the hope that this generalizes to the test data as well.  The amount of the changes depends on the contribution of a data point in the loss.  So, if a training data point leads to a large loss, the SGD algorithm will influence some parameters to adapt themselves towards reducing the loss on this point.  If the data point is not seen by the model during training, the changes in the gradient on this point is gradual throughout the training.  This is what we exploit in our active membership inference attack. 

Let $\vec{x}$ be a data record, which is targeted by the adversary to determine its membership.  Let us assume the adversary is one of the participants.  The attacker runs a gradient {\em ascent} on $\vec{x}$, and updates its local model parameters in the direction of increasing the loss on $\vec{x}$.  This can simply be done by adding the gradient to the parameters, 
\begin{align}
	\vec{W} \leftarrow \vec{W} + \gamma \frac{\partial L^{\vec{x}}}{\partial \vec{W}},
\end{align} 
where $\gamma$ is the adversarial update rate. The adversary then uploads the adversarially computed parameters to the central server, who will aggregate them with the parameter updates from other participants.  The adversary can run this attack on a batch of target data points all at the same time.

If the target record $\vec{x}$ is in the training set of a participant, its local SGD algorithm abruptly reduces the gradient of the loss on $\vec{x}$.  This can be detected by the inference model, and be used to distinguish members from non-members.  Repeated active attacks, which happens in federated learning, lead to high confidence inference attacks.

\subsection{Prior Knowledge: Supervised vs.\ Unsupervised Inference}

\label{sec:inference_supervised_unsupervised}

To construct his inference attack model, the adversary needs to find the meaningful mapping between the model's behavior on a data point and its membership in the training set.  The most straightforward way of learning such relationship is through some known members of the training data, and some data points from the same distribution which are not in the training data set.  This is illustrated in Table~\ref{tab:framework}.  The adversary has a dataset $\set{D}'$ that overlaps with the target dataset $\set{D}$.  Given this dataset, he can train the attack model in a supervised way, and use it to attack the rest of the training dataset.

Let $h$ be the inference attack model.  In the supervised setting, we minimize the (mean square) loss of the attacker for predicting the membership of the data points in its training set $\set{D}'$:
\begin{align}
	\sum_{d \in \set{D}'\cap\set{D}}  (h(d)-1)^2 + \sum_{d \in \set{D}'\setminus\set{D}}  (h(d))^2
\end{align}

If the adversary does not have known samples from the target training set, there are two possibilities for training the inference attack models: supervised training on shadow models~\cite{shokri2017membership}, and unsupervised training on the target model.  Shadow models are models with the same architecture as the target model.  The training data records for the shadow models are generated from the same distribution as the target training data, but do not have a known overlap with the target training set.  The attacker trains the attack model on the shadow models.  As the behavior of the shadow models on their training data is more or less similar to the behavior of the target model on its training data, the attack models trained on the shadow models are empirically shown to be effective. 

The attack output for (shadow) supervised training setting is the probability of membership.
\begin{align}
	h(d) = \Pr(d \in \set{D}; f)
\end{align}

\paragraphbe{Unsupervised training of inference models.}  We introduce an alternative approach to shadow training, which is unsupervised training of the attack model on the target model.  The assumption for this attack is that the attacker has access to a dataset $\set{D}'$ which partially overlaps with the target training set $\set{D}$, however, the adversary does not know which data points are in $\set{D}'\cap\set{D}$.

Our objective is to find a score for each data point that represents its  embedding in a space, which helps us easily separating members from non-members (using clustering algorithms).  The attack's output should compute such representations.  We make use of an encoder-decoder architecture to achieve this.  This is very similar to the auto-encoders for unsupervised deep learning.  As shown in Figure~\ref{fig:attack_model}, the output of the attack is fed into a decoder.  The decoder is a fully connected network with one hidden layer.  

The objective of the decoder is to reconstruct some key features of the attack input which are important for membership inference.  These include the loss value $L$, whether the target model has predicted the correct label $\mathbbm{1}_{y = \arg\max f(\vec{x})}$, the confidence of the model on the correct label $f(\vec{x})_{y}$, the prediction uncertainty (entropy) of the model $\mathrm{H}(f(\vec{x}))$, and the norm of the gradients $\norm{\frac{\partial L}{\partial \vec{W}}}$.  As previous work~\cite{shokri2017membership} as well as our empirical results show, these features are strong signals for distinguishing members from non-members.  The encoder-decoder architecture maximizes the information that the attack output contains about these features.  Thus, it generates a membership embedding for each data point.  Note that after training the attack model, the decoder plays no role in the membership inference attack.  

The attack in the unsupervised setting is a batch attack, where the adversary attacks a large set of data records (disjoint from his background knowledge).  We will use the encoder to for each target data record, and we compute the embedding value (output of the encoder model). Next, we use a clustering algorithm (e.g., we use the spectral clustering method) to cluster each input of the target model in two clusters.  Note that the outcome of the clustering algorithm is a threshold, as the attack output is a single number.  We predict the cluster with the larger gradient norm as non-members. 

\section{Experimental Setup}

We implemented our attacks using Pytorch.\footnote{{https://pytorch.org/}} We trained all of the models on a PC equipped with four Titan X GPU each with $12$ GB of memory.

\subsection{Datasets}

We used three datasets in our experiments: a standard image recognition benchmark dataset, CIFAR100, and two datasets Purchase100 and Texas100~\cite{shokri2017membership}.

\paragraphbe{CIFAR100.} 
This is a popular benchmark dataset used to evaluate image recognition algorithms~\cite{krizhevsky2009learning}.  It contains $60,000$ color (RGB) images, each   $32\times32$ pixels. The images are clustered into $100$ classes based on objects in the images.

\paragraphbe{Purchase100.}
The Purchase100 dataset contains the shopping records of several thousand online customers,  extracted during  Kaggle's ``acquire valued shopper'' challenge.\footnote{{https://www.kaggle.com/c/acquire-valued-shoppers-challenge/data}} 
The challenge was designed to identify offers that would attract new shoppers.  
We used the processed and simplified version of this dataset (courtesy of the authors of \cite{shokri2017membership}). Each record in the dataset is the shopping history of a single user. 
The dataset contains $600$ different products, and each user has a binary record which indicates whether she has bought each of the products (a total of  $197,324$ data records). The records are clustered into $100$ classes based on the similarity of the purchases, and our objective is to identify the class of each user's purchases. 

\paragraphbe{Texas100.}
This dataset includes hospital discharge data records released by the Texas Department of State Health Services~\footnote{{https://www.dshs.texas.gov/thcic/hospitals/Inpatientpudf.shtm}}. 
The records contain generic information about the patients (gender, age, and race), external causes of injury (e.g., drug misuse), the diagnosis, and patient procedures. Similar to Purchase100, we obtained the processed dataset (Courtesy of the authors~\cite{shokri2017membership}), which contains $67,330$ records and $6,170$ binary features.

\subsection{Target Models}
We investigate our attack model on the previously mentioned three datasets, Texas100, Purchase100 and CIFAR100. For the CIFAR100 dataset we used Alexnet~\cite{krizhevsky2012imagenet}, ResNet~\cite{he2016deep}, DenseNet~\cite{huang2017densely} models. We used SGD optimizer~\cite{goodfellow2016deep} to train the CIFAR100 models with learning rates of $0.01, 0.001, 0.0001$ for epochs $0-50, 50-100, 100-300$ accordingly. We used $l2$ regularization with weight of $0.0005$. 

For the Texas100 and Purchase100 datasets, we used fully connected models. For Purchase100, we used a model with layer sizes of ${600,1024,512,256,128,100}$ (where 100 is the output layer), and for Texas100, we used layers with size ${1024,512,256,128,100}$ (where 100 is the output layer). We used Adam~\cite{goodfellow2016deep} optimizer with the learning rate of $0.001$ for learning of these models. We trained each model for $100$ epochs across all of our experiments. We selected the model with the best testing accuracy across all the $100$ epochs.

\subsection{Pre-trained Models}

To demonstrate that our attacks are not limited to our training algorithm, we used publicly available pre-trained CIFAR100 models\footnote{We make use of ResNet, DenseNet, and Alexnet pre-trained models, provided in  {https://github.com/bearpaw/pytorch-classification}}. 
 All of these models are tuned to get the best testing accuracy using different regularization techniques. 

\subsection{Federated Learning}

We performed the training for all of the federated learning experiments. Specifically, we used the averaging aggregation method for the federated scenario~\cite{konevcny2016federated}. Each training party sends the parameter updates after every epoch of training to the central model, and the central server averages the models' updates from the parties and sends the updated model to all parties. In our experiments, we use the same training dataset size for all parties, and each party's training data is selected uniformly at random from our available datasets. 

\subsection{Attack Models}
Table~\ref{tab:model_sizes} in Appendix~\ref{sec:appendix}, presents the details of  our attack model architecture. As can be seen, we used  ReLU activation functions, and we initialized the weights using a normal distribution with mean $0$ and standard deviation of $0.01$. The bias values of all layers are initialized with $0$. The batch size of all experiments is $64$. To train the attack model we use the Adam optimizer with a learning rate of $0.0001$. We train attack models for $100$ epochs and pick the model with the highest testing accuracy, across all the $100$ epochs. 

Tables~\ref{tab:datasize} and \ref{tab:datasize_fed} present the dataset sizes used for training the target and attack models.  In the supervised setting for training the attack models, we assume the attacker has access to a fraction of the training set and some non-member samples.  In this case, to balance the training, we select half of each batch to include member instances and the other half non-member instances from the attacker's background knowledge. Creating the batches in this fashion will prevent the attack model from a bias towards member or non-member instances.

\subsection{Evaluation Metrics}

\paragraphbe{Attack accuracy}
The attacker's output has two classes ``\textit{Member}'' and ``\textit{Non-member}''. Attack accuracy is the fraction of the correct membership predictions (predicting members as member and non-members as non-member) for unknown data points.  The size of the set of member and non-member samples that we use for evaluating the attack are the same. 

\paragraphbe{True/False positive} For a more detailed evaluation of attack performance, we also measure the true positive and false positive rates of the attacker. Positive is associated with the attacker outputting ``member''.

\paragraphbe{Prediction uncertainty}
For a classification model, we compute its prediction uncertainty using the normalized entropy of its prediction vector for a given input.
\begin{align}
\text{H} = \frac{1}{\log(K)} \sum_{i=1}^{K}p_i \log(p_i)
\end{align}
where  $K$ is the number of all classes and $p_i$ is the prediction probability for the  $i$th class. 
We compute the probabilities using a softmax function as $p_i=\frac{e^{h(d)^{(i)}}}{\sum_{k=1}^{K}e^{h(d)^{(k)}}}$.

\begin{table}
    \centering
        \caption{{\small Size of datasets used for training and testing the target classification model and the membership inference model}}
    \label{tab:datasize}

    \resizebox{\columnwidth}{!}{%
    \begin{tabular}{|c||c|c||c|c|c|c|}
        \hline
        & \multicolumn{2}{ c|| }{\textbf{Target Model}} &  \multicolumn{4}{ c| }{\textbf{Inference Attack Model}} \\ \cline{2-7}
        \multirow{2}{*}{\textbf{Datasets}} & \multirow{2}{*}{\textbf{Training}}   & \multirow{2}{*}{\textbf{Test}}  & \textbf{Training} & \textbf{Training} &\textbf{Test } & \textbf{Test}  \\
        &&&\textbf{Members} &\textbf{Non-members}&\textbf{Members} &\textbf{Non-members}\\
        \hline
        CIFAR100 & 50,000 & 10,000& 25,000 & 5,000 & 5,000& 5,000\\  
        Texas100 & 10,000 & 70,000 & 5,000& 10,000 & 10,000& 10,000\\
        Puchase100 & 20,000 & 50,000&10,000&10,000&10,000&10,000 \\
        \hline
    \end{tabular}
    }
\end{table}

\section{Experiments}
\label{sec:experiments}

We start by presenting our results for the stand-alone scenario, followed by our results for the federated learning scenario. 

\subsection{Stand-Alone Setting: Attacking Fully-Trained Models}

We investigate the case where the attacker has access to the fully-trained target model, in the white-box setting. Therefore, the attacker can leverage the outputs and the gradients of the hidden layers of the target model to perform the attack.
We have used pre-trained CIFAR100 models, and have trained other target models and the attack models using the dataset sizes which are presented in  Table~\ref{tab:datasize}.

\paragraphbe{Impact of the outputs of different layers:}
To understand and demonstrate the impact of different layers' outputs, we perform the attack separately using the outputs of \emph{individual layers}. 
We use a pre-trained Alexnet model as the target model, where the model is composed of five convolutional layers and a fully connected layer at the end. 
Table~\ref{tab:cifar_diff_layer} shows the accuracy of the attack using the output of each of the last three layers. 
As the table shows, using the last layers results in the highest attack accuracy, i.e., \textbf{among the layer outputs, the last layer (model output) leaks the most membership information about the training data}.The reason behind this is twofold.
By proceeding to the later layers, the capacity of the parameters ramps up, which leads the target model to store unnecessary information about the training dataset, and therefore leak more information.  Moreover, the first layers extract simple features from the input, thus generalize much better compared to the last layers, which are responsible for complex task of finding the relationship between abstract features and the classes.  We did not achieve significant accuracy gain by combining the outputs from multiple layers; this is because the leakage from the last layer (which is equivalent to a black-box inference attack) already contains the membership information that leaks from the output of the previous layers.

\paragraphbe{Impact of gradients:}
In Section~\ref{sec:inference_blackbox_whitebox}, we discussed why gradients should  leak information about the training dataset. In Table~\ref{tab:best_white_models}, we compare the accuracy of the membership attack when the attacker uses the gradients versus layer outputs, for different dataset and models. 
Overall, the results show that \emph{gradients leak significantly more membership information about the training set, compared to the layer outputs}. 

We compare the result of the attack on pre-trained CIFAR100-ResNet and CIFAR100-DenseNet models, where both are designed for the same image recognition task,  both are trained on the same dataset, and both have similar generalization error. The results show that these two models have various degrees of membership leakage; this suggests that the \textbf{generalization error is not the right metric to quantify privacy leakage in the white-box setting. The large capacity of the model which enables it to learn complex tasks and generalize well, leads to also memorizing more information about the training data.} The total number of the parameters in pre-trained Densenet model is $25.62$M , whereas this is only $1.7$M parameters for ResNet.

We also investigated the impact of gradients of different layers on attack accuracy. 
The results are shown in Table~\ref{tab:cifar_diff_grads}  show that  \textbf{the gradient of the later layers leak more membership information}.  This is similar to our findings for layer outputs: the last layer generalizes the least among all the layers in the model, and is the most dependent layer to the training set. 
By combining the gradients of all layers, we are able to only slightly increase the attack accuracy. 

Finally, Table~\ref{tab:cifar_dif_out_grad}  shows the attack accuracy when we combine the output layer and gradients of different layers.  We see that the gradients from the last layer leak the most membership information.

\paragraphbe{Impact of the training size:} 
Table~\ref{tab:diff_data_size} shows attack accuracy for various sizes of the attacker's  training data. 
The models are tested on the same set of test instances, across all these scenarios. 
As expected, increasing the size of the attacker's training dataset improves the accuracy of the membership inference attack. 

\begin{figure}
\centering
\begin{tikzpicture}
    \tikzstyle{every node}=[font=\small]

\begin{axis}[
height=\figureheight,
legend cell align={left},
legend entries={{Member instances},{Non-member instances}},
legend style={draw=white!80.0!black},
tick align=outside,
tick pos=left,
width=\figurewidth,
y label style={at={(axis description cs:0.1,.5)},anchor=south},
x grid style={lightgray!92.02614379084967!black},
xlabel={Training epoch},
xmajorgrids,
xmin=-4.75, xmax=99.75,
y grid style={lightgray!92.02614379084967!black},
ylabel={Gradient norm},
ymajorgrids,
ytick={0,0.1,...,1},
ymin=-0.0351438404351423, ymax=0.738022207651961
]
\addplot [line width=0.5599999999999999pt, black, mark=square*, mark size=1.5, mark options={solid}]
table [row sep=\\]{%
0	0.669249653816223 \\
5	0.336214572191238 \\
10	0.136452779173851 \\
15	0.0358518287539482 \\
20	0.0194540396332741 \\
25	0.00570545252412558 \\
30	0.025003906339407 \\
35	0.0230141989886761 \\
40	0.0116346534341574 \\
45	0.00244571827352047 \\
50	8.2866259617731e-05 \\
55	8.42470672068885e-06 \\
60	3.49877450389613e-06 \\
65	1.78730965672003e-06 \\
70	9.9405281162035e-07 \\
75	5.9618497516567e-07 \\
80	3.46766853454028e-07 \\
85	2.08684696190176e-07 \\
90	1.21849112133532e-07 \\
95	7.08415441863508e-08 \\
};
\addplot [line width=0.5599999999999999pt, black,mark=x, mark size=1.5, mark options={solid}]
table [row sep=\\]{%
0	0.702878296375275 \\
5	0.374176025390625 \\
10	0.263767182826996 \\
15	0.221414729952812 \\
20	0.275614082813263 \\
25	0.222538039088249 \\
30	0.212232753634453 \\
35	0.19686621427536 \\
40	0.213182508945465 \\
45	0.248033344745636 \\
50	0.238106787204742 \\
55	0.221253767609596 \\
60	0.217506676912308 \\
65	0.217142909765244 \\
70	0.216512829065323 \\
75	0.214945957064629 \\
80	0.21495446562767 \\
85	0.215371072292328 \\
90	0.214653447270393 \\
95	0.213721871376038 \\
};
\end{axis}
\end{tikzpicture}
    \caption{\small Gradient norms of the last layer during  learning epochs for member and non-member instances (for Purchase100). 
    }
    \label{fig:grad_norm_train_test_purchase100}
\end{figure}
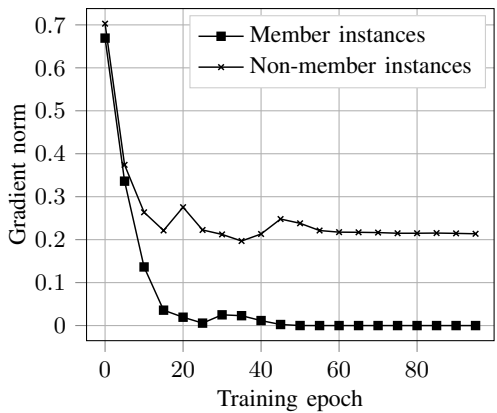

\paragraphbe{Impact of the gradient norm:} 
In this experiment, we demonstrate that the norm of the model's gradients is highly correlated with the accuracy of membership inference, as it behaves differently for member and non-member instances. 
Figure~\ref{fig:grad_norm_train_test_purchase100} plots the last-layer gradient norms over consecutive training epochs for member and non-member instances (for the Purchase100 dataset). 
As can be seen, during training, the gradient norms of the member instances decrease over training epochs, which is not the case for non-member instances.

\begin{figure*}
    \begin{subfigure}{0.33\linewidth}
    \centering
\begin{tikzpicture}
    \tikzstyle{every node}=[font=\small]

\begin{axis}[
height=\trifigureheight,
legend cell align={left},
legend entries={{Member},{Non-member}},
legend style={draw=white!80.0!black},
tick align=outside,
tick pos=left,
width=\trifigurewidth,
y label style={at={(axis description cs:0.1,.5)},anchor=south},
x grid style={lightgray!92.02614379084967!black},
xlabel={Gradient norm},
xmajorgrids,
xmin=-100, xmax=3000,
y grid style={lightgray!92.02614379084967!black},
ylabel={Fraction},
x label style={at={(axis description cs:0.5,-.25)},anchor=south},
xticklabel style={rotate=45},
ymajorgrids,
ymin=0, ymax=0.97125000000038,
ytick={0,0.2,0.4,0.6,0.8,1},
xtick={0,500,...,3000},
yticklabels={0.0,0.2,0.4,0.6,0.8,1.0}
]
\addlegendimage{ybar,ybar legend,draw=black,fill=white!82.74509803921568!black};
\draw[draw=black,fill=white!82.74509803921568!black] (axis cs:-100,0) rectangle (axis cs:0,0.925000000000362);
\draw[draw=black,fill=white!82.74509803921568!black] (axis cs:348.275862068966,0) rectangle (axis cs:448.275862068966,0.0480399999999987);
\draw[draw=black,fill=white!82.74509803921568!black] (axis cs:796.551724137931,0) rectangle (axis cs:896.551724137931,0.01576);
\draw[draw=black,fill=white!82.74509803921568!black] (axis cs:1244.8275862069,0) rectangle (axis cs:1344.8275862069,0.00600000000000001);
\draw[draw=black,fill=white!82.74509803921568!black] (axis cs:1693.10344827586,0) rectangle (axis cs:1793.10344827586,0.00276);
\draw[draw=black,fill=white!82.74509803921568!black] (axis cs:2141.37931034483,0) rectangle (axis cs:2241.37931034483,0.00064);
\draw[draw=black,fill=white!82.74509803921568!black] (axis cs:2589.65517241379,0) rectangle (axis cs:2689.65517241379,0.0008);
\draw[draw=black,fill=white!82.74509803921568!black] (axis cs:3037.93103448276,0) rectangle (axis cs:3137.93103448276,0.00044);
\draw[draw=black,fill=white!82.74509803921568!black] (axis cs:3486.20689655172,0) rectangle (axis cs:3586.20689655172,0.00032);
\draw[draw=black,fill=white!82.74509803921568!black] (axis cs:3934.48275862069,0) rectangle (axis cs:4034.48275862069,0.00012);
\draw[draw=black,fill=white!82.74509803921568!black] (axis cs:4382.75862068965,0) rectangle (axis cs:4482.75862068965,0);
\draw[draw=black,fill=white!82.74509803921568!black] (axis cs:4831.03448275862,0) rectangle (axis cs:4931.03448275862,0);
\draw[draw=black,fill=white!82.74509803921568!black] (axis cs:5279.31034482759,0) rectangle (axis cs:5379.31034482759,0);
\draw[draw=black,fill=white!82.74509803921568!black] (axis cs:5727.58620689655,0) rectangle (axis cs:5827.58620689655,0);
\draw[draw=black,fill=white!82.74509803921568!black] (axis cs:6175.86206896552,0) rectangle (axis cs:6275.86206896552,0);
\draw[draw=black,fill=white!82.74509803921568!black] (axis cs:6624.13793103448,0) rectangle (axis cs:6724.13793103448,4e-05);
\draw[draw=black,fill=white!82.74509803921568!black] (axis cs:7072.41379310345,0) rectangle (axis cs:7172.41379310345,0);
\draw[draw=black,fill=white!82.74509803921568!black] (axis cs:7520.68965517241,0) rectangle (axis cs:7620.68965517241,0);
\draw[draw=black,fill=white!82.74509803921568!black] (axis cs:7968.96551724138,0) rectangle (axis cs:8068.96551724138,4e-05);
\draw[draw=black,fill=white!82.74509803921568!black] (axis cs:8417.24137931034,0) rectangle (axis cs:8517.24137931034,0);
\draw[draw=black,fill=white!82.74509803921568!black] (axis cs:8865.51724137931,0) rectangle (axis cs:8965.51724137931,0);
\draw[draw=black,fill=white!82.74509803921568!black] (axis cs:9313.79310344828,0) rectangle (axis cs:9413.79310344828,0);
\draw[draw=black,fill=white!82.74509803921568!black] (axis cs:9762.06896551724,0) rectangle (axis cs:9862.06896551724,4e-05);
\draw[draw=black,fill=white!82.74509803921568!black] (axis cs:10210.3448275862,0) rectangle (axis cs:10310.3448275862,0);
\draw[draw=black,fill=white!82.74509803921568!black] (axis cs:10658.6206896552,0) rectangle (axis cs:10758.6206896552,0);
\draw[draw=black,fill=white!82.74509803921568!black] (axis cs:11106.8965517241,0) rectangle (axis cs:11206.8965517241,0);
\draw[draw=black,fill=white!82.74509803921568!black] (axis cs:11555.1724137931,0) rectangle (axis cs:11655.1724137931,0);
\draw[draw=black,fill=white!82.74509803921568!black] (axis cs:12003.4482758621,0) rectangle (axis cs:12103.4482758621,0);
\draw[draw=black,fill=white!82.74509803921568!black] (axis cs:12451.724137931,0) rectangle (axis cs:12551.724137931,0);
\draw[draw=black,fill=white!82.74509803921568!black] (axis cs:12900,0) rectangle (axis cs:13000,0);
\addlegendimage{ybar,ybar legend,draw=white,fill=black};
\draw[draw=white,fill=black] (axis cs:0,0) rectangle (axis cs:100,0.435199999999985);
\draw[draw=white,fill=black] (axis cs:448.275862068966,0) rectangle (axis cs:548.275862068966,0.142400000000002);
\draw[draw=white,fill=black] (axis cs:896.551724137931,0) rectangle (axis cs:996.551724137931,0.126800000000002);
\draw[draw=white,fill=black] (axis cs:1344.8275862069,0) rectangle (axis cs:1444.8275862069,0.0896000000000005);
\draw[draw=white,fill=black] (axis cs:1793.10344827586,0) rectangle (axis cs:1893.10344827586,0.0623999999999997);
\draw[draw=white,fill=black] (axis cs:2241.37931034483,0) rectangle (axis cs:2341.37931034483,0.0433999999999998);
\draw[draw=white,fill=black] (axis cs:2689.65517241379,0) rectangle (axis cs:2789.65517241379,0.0325999999999999);
\draw[draw=white,fill=black] (axis cs:3137.93103448276,0) rectangle (axis cs:3237.93103448276,0.02);
\draw[draw=white,fill=black] (axis cs:3586.20689655172,0) rectangle (axis cs:3686.20689655172,0.0142);
\draw[draw=white,fill=black] (axis cs:4034.48275862069,0) rectangle (axis cs:4134.48275862069,0.00779999999999999);
\draw[draw=white,fill=black] (axis cs:4482.75862068965,0) rectangle (axis cs:4582.75862068965,0.0068);
\draw[draw=white,fill=black] (axis cs:4931.03448275862,0) rectangle (axis cs:5031.03448275862,0.005);
\draw[draw=white,fill=black] (axis cs:5379.31034482759,0) rectangle (axis cs:5479.31034482759,0.0032);
\draw[draw=white,fill=black] (axis cs:5827.58620689655,0) rectangle (axis cs:5927.58620689655,0.003);
\draw[draw=white,fill=black] (axis cs:6275.86206896552,0) rectangle (axis cs:6375.86206896552,0.0024);
\draw[draw=white,fill=black] (axis cs:6724.13793103448,0) rectangle (axis cs:6824.13793103448,0.0002);
\draw[draw=white,fill=black] (axis cs:7172.41379310345,0) rectangle (axis cs:7272.41379310345,0.001);
\draw[draw=white,fill=black] (axis cs:7620.68965517241,0) rectangle (axis cs:7720.68965517241,0.0012);
\draw[draw=white,fill=black] (axis cs:8068.96551724138,0) rectangle (axis cs:8168.96551724138,0.0004);
\draw[draw=white,fill=black] (axis cs:8517.24137931034,0) rectangle (axis cs:8617.24137931034,0.0006);
\draw[draw=white,fill=black] (axis cs:8965.51724137931,0) rectangle (axis cs:9065.51724137931,0.0002);
\draw[draw=white,fill=black] (axis cs:9413.79310344828,0) rectangle (axis cs:9513.79310344828,0.0004);
\draw[draw=white,fill=black] (axis cs:9862.06896551724,0) rectangle (axis cs:9962.06896551724,0);
\draw[draw=white,fill=black] (axis cs:10310.3448275862,0) rectangle (axis cs:10410.3448275862,0.0004);
\draw[draw=white,fill=black] (axis cs:10758.6206896552,0) rectangle (axis cs:10858.6206896552,0.0004);
\draw[draw=white,fill=black] (axis cs:11206.8965517241,0) rectangle (axis cs:11306.8965517241,0.0002);
\draw[draw=white,fill=black] (axis cs:11655.1724137931,0) rectangle (axis cs:11755.1724137931,0);
\draw[draw=white,fill=black] (axis cs:12103.4482758621,0) rectangle (axis cs:12203.4482758621,0);
\draw[draw=white,fill=black] (axis cs:12551.724137931,0) rectangle (axis cs:12651.724137931,0.0002);
\draw[draw=white,fill=black] (axis cs:13000,0) rectangle (axis cs:13100,0);
\end{axis}

\end{tikzpicture}
            \caption{CIFAR100-Alexnet}
        \end{subfigure}
        \begin{subfigure}{0.33\linewidth}
    \centering
\begin{tikzpicture}
    \tikzstyle{every node}=[font=\small]

\begin{axis}[
height=\trifigureheight,
legend cell align={left},
legend entries={{Member},{Non-member}},
legend style={draw=white!80.0!black},
tick align=outside,
tick pos=left,
width=\trifigurewidth,
y label style={at={(axis description cs:0.1,.5)},anchor=south},
x grid style={lightgray!92.02614379084967!black},
xlabel={Gradient norm},
xmajorgrids,
xmin=-10, xmax=300,
x label style={at={(axis description cs:0.5,-.25)},anchor=south},
xticklabel style={rotate=45},
y grid style={lightgray!92.02614379084967!black},
ylabel={Fraction},
ymajorgrids,
ymin=0, ymax=1.04947499999994,
ytick={0,0.2,0.4,0.6,0.8,1,1.2},
xtick={0,50,...,300},
yticklabels={0.0,0.2,0.4,0.6,0.8,1.0,1.2}
]
\addlegendimage{ybar,ybar legend,draw=black,fill=white!82.74509803921568!black};
\draw[draw=black,fill=white!82.74509803921568!black] (axis cs:-10,0) rectangle (axis cs:0,0.999499999999945);
\draw[draw=black,fill=white!82.74509803921568!black] (axis cs:17.5862068965517,0) rectangle (axis cs:27.5862068965517,0.0005);
\draw[draw=black,fill=white!82.74509803921568!black] (axis cs:45.1724137931034,0) rectangle (axis cs:55.1724137931034,0);
\draw[draw=black,fill=white!82.74509803921568!black] (axis cs:72.7586206896552,0) rectangle (axis cs:82.7586206896552,0);
\draw[draw=black,fill=white!82.74509803921568!black] (axis cs:100.344827586207,0) rectangle (axis cs:110.344827586207,0);
\draw[draw=black,fill=white!82.74509803921568!black] (axis cs:127.931034482759,0) rectangle (axis cs:137.931034482759,0);
\draw[draw=black,fill=white!82.74509803921568!black] (axis cs:155.51724137931,0) rectangle (axis cs:165.51724137931,0);
\draw[draw=black,fill=white!82.74509803921568!black] (axis cs:183.103448275862,0) rectangle (axis cs:193.103448275862,0);
\draw[draw=black,fill=white!82.74509803921568!black] (axis cs:210.689655172414,0) rectangle (axis cs:220.689655172414,0);
\draw[draw=black,fill=white!82.74509803921568!black] (axis cs:238.275862068965,0) rectangle (axis cs:248.275862068965,0);
\draw[draw=black,fill=white!82.74509803921568!black] (axis cs:265.862068965517,0) rectangle (axis cs:275.862068965517,0);
\draw[draw=black,fill=white!82.74509803921568!black] (axis cs:293.448275862069,0) rectangle (axis cs:303.448275862069,0);
\draw[draw=black,fill=white!82.74509803921568!black] (axis cs:321.034482758621,0) rectangle (axis cs:331.034482758621,0);
\draw[draw=black,fill=white!82.74509803921568!black] (axis cs:348.620689655172,0) rectangle (axis cs:358.620689655172,0);
\draw[draw=black,fill=white!82.74509803921568!black] (axis cs:376.206896551724,0) rectangle (axis cs:386.206896551724,0);
\draw[draw=black,fill=white!82.74509803921568!black] (axis cs:403.793103448276,0) rectangle (axis cs:413.793103448276,0);
\draw[draw=black,fill=white!82.74509803921568!black] (axis cs:431.379310344828,0) rectangle (axis cs:441.379310344828,0);
\draw[draw=black,fill=white!82.74509803921568!black] (axis cs:458.965517241379,0) rectangle (axis cs:468.965517241379,0);
\draw[draw=black,fill=white!82.74509803921568!black] (axis cs:486.551724137931,0) rectangle (axis cs:496.551724137931,0);
\draw[draw=black,fill=white!82.74509803921568!black] (axis cs:514.137931034483,0) rectangle (axis cs:524.137931034483,0);
\draw[draw=black,fill=white!82.74509803921568!black] (axis cs:541.724137931034,0) rectangle (axis cs:551.724137931034,0);
\draw[draw=black,fill=white!82.74509803921568!black] (axis cs:569.310344827586,0) rectangle (axis cs:579.310344827586,0);
\draw[draw=black,fill=white!82.74509803921568!black] (axis cs:596.896551724138,0) rectangle (axis cs:606.896551724138,0);
\draw[draw=black,fill=white!82.74509803921568!black] (axis cs:624.48275862069,0) rectangle (axis cs:634.48275862069,0);
\draw[draw=black,fill=white!82.74509803921568!black] (axis cs:652.068965517241,0) rectangle (axis cs:662.068965517241,0);
\draw[draw=black,fill=white!82.74509803921568!black] (axis cs:679.655172413793,0) rectangle (axis cs:689.655172413793,0);
\draw[draw=black,fill=white!82.74509803921568!black] (axis cs:707.241379310345,0) rectangle (axis cs:717.241379310345,0);
\draw[draw=black,fill=white!82.74509803921568!black] (axis cs:734.827586206897,0) rectangle (axis cs:744.827586206897,0);
\draw[draw=black,fill=white!82.74509803921568!black] (axis cs:762.413793103448,0) rectangle (axis cs:772.413793103448,0);
\draw[draw=black,fill=white!82.74509803921568!black] (axis cs:790,0) rectangle (axis cs:800,0);
\addlegendimage{ybar,ybar legend,draw=white,fill=black};
\draw[draw=white,fill=black] (axis cs:0,0) rectangle (axis cs:10,0.786999999999969);
\draw[draw=white,fill=black] (axis cs:27.5862068965517,0) rectangle (axis cs:37.5862068965517,0.0235);
\draw[draw=white,fill=black] (axis cs:55.1724137931034,0) rectangle (axis cs:65.1724137931034,0.018);
\draw[draw=white,fill=black] (axis cs:82.7586206896552,0) rectangle (axis cs:92.7586206896552,0.019);
\draw[draw=white,fill=black] (axis cs:110.344827586207,0) rectangle (axis cs:120.344827586207,0.0245);
\draw[draw=white,fill=black] (axis cs:137.931034482759,0) rectangle (axis cs:147.931034482759,0.023);
\draw[draw=white,fill=black] (axis cs:165.51724137931,0) rectangle (axis cs:175.51724137931,0.013);
\draw[draw=white,fill=black] (axis cs:193.103448275862,0) rectangle (axis cs:203.103448275862,0.018);
\draw[draw=white,fill=black] (axis cs:220.689655172414,0) rectangle (axis cs:230.689655172414,0.0165);
\draw[draw=white,fill=black] (axis cs:248.275862068965,0) rectangle (axis cs:258.275862068965,0.0145);
\draw[draw=white,fill=black] (axis cs:275.862068965517,0) rectangle (axis cs:285.862068965517,0.0115);
\draw[draw=white,fill=black] (axis cs:303.448275862069,0) rectangle (axis cs:313.448275862069,0.011);
\draw[draw=white,fill=black] (axis cs:331.034482758621,0) rectangle (axis cs:341.034482758621,0.0085);
\draw[draw=white,fill=black] (axis cs:358.620689655172,0) rectangle (axis cs:368.620689655172,0.004);
\draw[draw=white,fill=black] (axis cs:386.206896551724,0) rectangle (axis cs:396.206896551724,0.003);
\draw[draw=white,fill=black] (axis cs:413.793103448276,0) rectangle (axis cs:423.793103448276,0.0005);
\draw[draw=white,fill=black] (axis cs:441.379310344828,0) rectangle (axis cs:451.379310344828,0.0015);
\draw[draw=white,fill=black] (axis cs:468.965517241379,0) rectangle (axis cs:478.965517241379,0.0015);
\draw[draw=white,fill=black] (axis cs:496.551724137931,0) rectangle (axis cs:506.551724137931,0);
\draw[draw=white,fill=black] (axis cs:524.137931034483,0) rectangle (axis cs:534.137931034483,0);
\draw[draw=white,fill=black] (axis cs:551.724137931034,0) rectangle (axis cs:561.724137931034,0.0005);
\draw[draw=white,fill=black] (axis cs:579.310344827586,0) rectangle (axis cs:589.310344827586,0);
\draw[draw=white,fill=black] (axis cs:606.896551724138,0) rectangle (axis cs:616.896551724138,0);
\draw[draw=white,fill=black] (axis cs:634.48275862069,0) rectangle (axis cs:644.48275862069,0);
\draw[draw=white,fill=black] (axis cs:662.068965517241,0) rectangle (axis cs:672.068965517241,0.0005);
\draw[draw=white,fill=black] (axis cs:689.655172413793,0) rectangle (axis cs:699.655172413793,0);
\draw[draw=white,fill=black] (axis cs:717.241379310345,0) rectangle (axis cs:727.241379310345,0);
\draw[draw=white,fill=black] (axis cs:744.827586206897,0) rectangle (axis cs:754.827586206897,0);
\draw[draw=white,fill=black] (axis cs:772.413793103448,0) rectangle (axis cs:782.413793103448,0);
\draw[draw=white,fill=black] (axis cs:800,0) rectangle (axis cs:810,0.0005);
\end{axis}

\end{tikzpicture}
            \caption{CIFAR100-Densenet}
            \label{fig:dense_grad}
        \end{subfigure}
        \begin{subfigure}{0.33\linewidth}
    \centering
\begin{tikzpicture}
    \tikzstyle{every node}=[font=\small]

\begin{axis}[
height=\trifigureheight,
legend cell align={left},
legend entries={{Member},{Non-member}},
legend style={draw=white!80.0!black},
tick align=outside,
tick pos=left,
y label style={at={(axis description cs:0.1,.5)},anchor=south},
width=\trifigurewidth,
x grid style={lightgray!92.02614379084967!black},
xlabel={Gradient norm},
xmajorgrids,
xmin=-10, xmax=300,
y grid style={lightgray!92.02614379084967!black},
ylabel={Fraction},
x label style={at={(axis description cs:0.5,-.25)},anchor=south},
xticklabel style={rotate=45},
ymajorgrids,
ymin=0, ymax=0.71032499999998,
ytick={0,0.2,0.4,0.6,0.8},
xtick={0,50,...,300},
yticklabels={0.0,0.2,0.4,0.6,0.8}
]
\addlegendimage{ybar,ybar legend,draw=black,fill=white!82.74509803921568!black};
\draw[draw=black,fill=white!82.74509803921568!black] (axis cs:-10,0) rectangle (axis cs:0,0.676499999999981);
\draw[draw=black,fill=white!82.74509803921568!black] (axis cs:17.5862068965517,0) rectangle (axis cs:27.5862068965517,0.1015);
\draw[draw=black,fill=white!82.74509803921568!black] (axis cs:45.1724137931034,0) rectangle (axis cs:55.1724137931034,0.0595);
\draw[draw=black,fill=white!82.74509803921568!black] (axis cs:72.7586206896552,0) rectangle (axis cs:82.7586206896552,0.04);
\draw[draw=black,fill=white!82.74509803921568!black] (axis cs:100.344827586207,0) rectangle (axis cs:110.344827586207,0.0355);
\draw[draw=black,fill=white!82.74509803921568!black] (axis cs:127.931034482759,0) rectangle (axis cs:137.931034482759,0.0235);
\draw[draw=black,fill=white!82.74509803921568!black] (axis cs:155.51724137931,0) rectangle (axis cs:165.51724137931,0.0185);
\draw[draw=black,fill=white!82.74509803921568!black] (axis cs:183.103448275862,0) rectangle (axis cs:193.103448275862,0.0145);
\draw[draw=black,fill=white!82.74509803921568!black] (axis cs:210.689655172414,0) rectangle (axis cs:220.689655172414,0.012);
\draw[draw=black,fill=white!82.74509803921568!black] (axis cs:238.275862068965,0) rectangle (axis cs:248.275862068965,0.0055);
\draw[draw=black,fill=white!82.74509803921568!black] (axis cs:265.862068965517,0) rectangle (axis cs:275.862068965517,0.0035);
\draw[draw=black,fill=white!82.74509803921568!black] (axis cs:293.448275862069,0) rectangle (axis cs:303.448275862069,0.0025);
\draw[draw=black,fill=white!82.74509803921568!black] (axis cs:321.034482758621,0) rectangle (axis cs:331.034482758621,0.0045);
\draw[draw=black,fill=white!82.74509803921568!black] (axis cs:348.620689655172,0) rectangle (axis cs:358.620689655172,0.001);
\draw[draw=black,fill=white!82.74509803921568!black] (axis cs:376.206896551724,0) rectangle (axis cs:386.206896551724,0.0005);
\draw[draw=black,fill=white!82.74509803921568!black] (axis cs:403.793103448276,0) rectangle (axis cs:413.793103448276,0.0005);
\draw[draw=black,fill=white!82.74509803921568!black] (axis cs:431.379310344828,0) rectangle (axis cs:441.379310344828,0);
\draw[draw=black,fill=white!82.74509803921568!black] (axis cs:458.965517241379,0) rectangle (axis cs:468.965517241379,0);
\draw[draw=black,fill=white!82.74509803921568!black] (axis cs:486.551724137931,0) rectangle (axis cs:496.551724137931,0);
\draw[draw=black,fill=white!82.74509803921568!black] (axis cs:514.137931034483,0) rectangle (axis cs:524.137931034483,0.0005);
\draw[draw=black,fill=white!82.74509803921568!black] (axis cs:541.724137931034,0) rectangle (axis cs:551.724137931034,0);
\draw[draw=black,fill=white!82.74509803921568!black] (axis cs:569.310344827586,0) rectangle (axis cs:579.310344827586,0);
\draw[draw=black,fill=white!82.74509803921568!black] (axis cs:596.896551724138,0) rectangle (axis cs:606.896551724138,0);
\draw[draw=black,fill=white!82.74509803921568!black] (axis cs:624.48275862069,0) rectangle (axis cs:634.48275862069,0);
\draw[draw=black,fill=white!82.74509803921568!black] (axis cs:652.068965517241,0) rectangle (axis cs:662.068965517241,0);
\draw[draw=black,fill=white!82.74509803921568!black] (axis cs:679.655172413793,0) rectangle (axis cs:689.655172413793,0);
\draw[draw=black,fill=white!82.74509803921568!black] (axis cs:707.241379310345,0) rectangle (axis cs:717.241379310345,0);
\draw[draw=black,fill=white!82.74509803921568!black] (axis cs:734.827586206897,0) rectangle (axis cs:744.827586206897,0);
\draw[draw=black,fill=white!82.74509803921568!black] (axis cs:762.413793103448,0) rectangle (axis cs:772.413793103448,0);
\draw[draw=black,fill=white!82.74509803921568!black] (axis cs:790,0) rectangle (axis cs:800,0);
\addlegendimage{ybar,ybar legend,draw=white,fill=black};
\draw[draw=white,fill=black] (axis cs:0,0) rectangle (axis cs:10,0.544999999999995);
\draw[draw=white,fill=black] (axis cs:27.5862068965517,0) rectangle (axis cs:37.5862068965517,0.0835000000000001);
\draw[draw=white,fill=black] (axis cs:55.1724137931034,0) rectangle (axis cs:65.1724137931034,0.0645);
\draw[draw=white,fill=black] (axis cs:82.7586206896552,0) rectangle (axis cs:92.7586206896552,0.052);
\draw[draw=white,fill=black] (axis cs:110.344827586207,0) rectangle (axis cs:120.344827586207,0.0455);
\draw[draw=white,fill=black] (axis cs:137.931034482759,0) rectangle (axis cs:147.931034482759,0.0445);
\draw[draw=white,fill=black] (axis cs:165.51724137931,0) rectangle (axis cs:175.51724137931,0.0425);
\draw[draw=white,fill=black] (axis cs:193.103448275862,0) rectangle (axis cs:203.103448275862,0.028);
\draw[draw=white,fill=black] (axis cs:220.689655172414,0) rectangle (axis cs:230.689655172414,0.031);
\draw[draw=white,fill=black] (axis cs:248.275862068965,0) rectangle (axis cs:258.275862068965,0.0185);
\draw[draw=white,fill=black] (axis cs:275.862068965517,0) rectangle (axis cs:285.862068965517,0.016);
\draw[draw=white,fill=black] (axis cs:303.448275862069,0) rectangle (axis cs:313.448275862069,0.013);
\draw[draw=white,fill=black] (axis cs:331.034482758621,0) rectangle (axis cs:341.034482758621,0.005);
\draw[draw=white,fill=black] (axis cs:358.620689655172,0) rectangle (axis cs:368.620689655172,0.0035);
\draw[draw=white,fill=black] (axis cs:386.206896551724,0) rectangle (axis cs:396.206896551724,0.002);
\draw[draw=white,fill=black] (axis cs:413.793103448276,0) rectangle (axis cs:423.793103448276,0.003);
\draw[draw=white,fill=black] (axis cs:441.379310344828,0) rectangle (axis cs:451.379310344828,0.001);
\draw[draw=white,fill=black] (axis cs:468.965517241379,0) rectangle (axis cs:478.965517241379,0.0015);
\draw[draw=white,fill=black] (axis cs:496.551724137931,0) rectangle (axis cs:506.551724137931,0);
\draw[draw=white,fill=black] (axis cs:524.137931034483,0) rectangle (axis cs:534.137931034483,0);
\draw[draw=white,fill=black] (axis cs:551.724137931034,0) rectangle (axis cs:561.724137931034,0);
\draw[draw=white,fill=black] (axis cs:579.310344827586,0) rectangle (axis cs:589.310344827586,0);
\draw[draw=white,fill=black] (axis cs:606.896551724138,0) rectangle (axis cs:616.896551724138,0);
\draw[draw=white,fill=black] (axis cs:634.48275862069,0) rectangle (axis cs:644.48275862069,0);
\draw[draw=white,fill=black] (axis cs:662.068965517241,0) rectangle (axis cs:672.068965517241,0);
\draw[draw=white,fill=black] (axis cs:689.655172413793,0) rectangle (axis cs:699.655172413793,0);
\draw[draw=white,fill=black] (axis cs:717.241379310345,0) rectangle (axis cs:727.241379310345,0);
\draw[draw=white,fill=black] (axis cs:744.827586206897,0) rectangle (axis cs:754.827586206897,0);
\draw[draw=white,fill=black] (axis cs:772.413793103448,0) rectangle (axis cs:782.413793103448,0);
\draw[draw=white,fill=black] (axis cs:800,0) rectangle (axis cs:810,0);
\end{axis}

\end{tikzpicture}
            \caption{CIFAR100-Resnet}
            \label{fig:resnet_grad}
        \end{subfigure}
        
    \caption{The distribution of gradient norms for  member and non-member instances of  different pretrained models.}
    \label{fig:grad_norms}
\end{figure*}
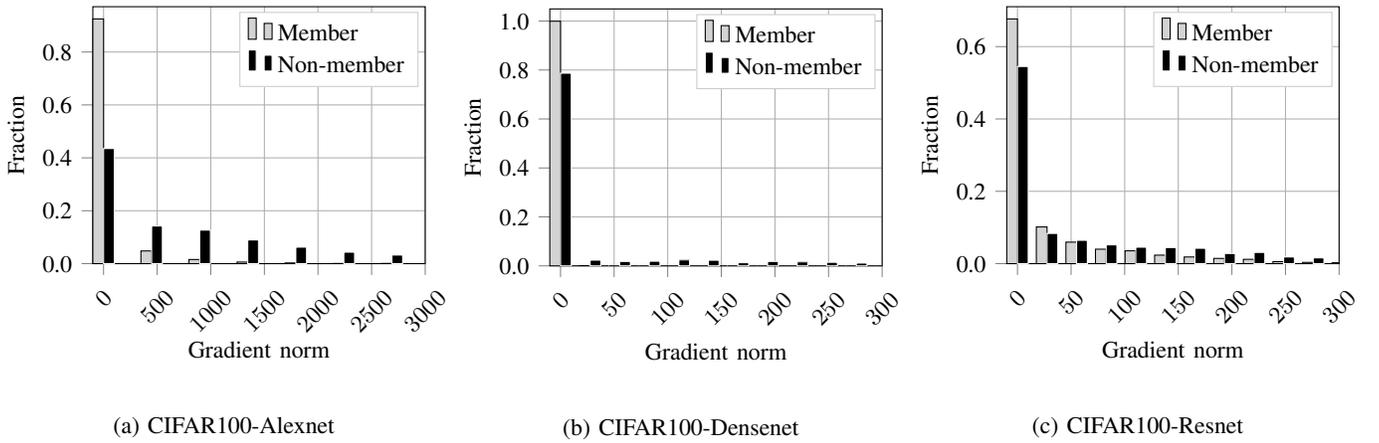

Figure~\ref{fig:grad_norms} shows the distribution of last-layer gradient norms for members and non-members on three various pretrained architectures on CIFAR100.
Comparing the figures with Table~\ref{tab:best_white_models}, we see that
 \emph{a model leaks more membership information when the distribution of the gradient norm is more distinct for member and non-member instances}. 
 For instance, we can see that ResNet and DenseNet both have relatively similar generalization errors, but the gradient norm distribution of members and non-members is more distinguishable for DenseNet (Figure~\ref{fig:dense_grad}) compared to ResNet (Figure~\ref{fig:resnet_grad}). We see that the attack accuracy in  DenseNet is much higher than ResNet.

\begin{figure}[t!]
    \centering
    \begin{subfigure}[t]{0.5\textwidth}
        \centering
\begin{tikzpicture}
    \tikzstyle{every node}=[font=\small]

\begin{axis}[
height=\figureheight,
legend cell align={left},
legend entries={{Member instances},{Non-member instances}},
legend style={at={(0.03,0.97)}, anchor=north west, draw=white!80.0!black},
tick align=outside,
tick pos=left,
width=\figurewidth,
x grid style={lightgray!92.02614379084967!black},
xlabel={Class},
xmajorgrids,
xmin=-4.95, xmax=103.95,
y grid style={lightgray!92.02614379084967!black},
ylabel={Gradient norm},
ymajorgrids,
ymin=-69.4933996200562, ymax=1963.93477916718
]
\addplot [line width=0.5599999999999999pt, black, mark=*, mark size=1, mark options={solid}]
table [row sep=\\]{%
0	54.5629806518555 \\
1	22.9351539611816 \\
2	88.3217544555664 \\
3	74.2037887573242 \\
4	42.865852355957 \\
5	66.1651153564453 \\
6	76.4579925537109 \\
7	91.5412139892578 \\
8	65.250846862793 \\
9	50.4485702514648 \\
10	24.497386932373 \\
11	189.402435302734 \\
12	44.8831977844238 \\
13	178.755142211914 \\
14	83.2888259887695 \\
15	75.0596389770508 \\
16	60.2619514465332 \\
17	126.432548522949 \\
18	99.1183929443359 \\
19	136.374450683594 \\
20	192.723754882812 \\
21	182.579345703125 \\
22	33.7406272888184 \\
23	96.0739517211914 \\
24	211.007797241211 \\
25	158.42317199707 \\
26	154.961669921875 \\
27	110.083549499512 \\
28	131.128265380859 \\
29	146.81494140625 \\
30	117.087799072266 \\
31	63.297191619873 \\
32	155.966583251953 \\
33	102.063201904297 \\
34	133.538284301758 \\
35	162.003753662109 \\
36	110.366424560547 \\
37	91.2087020874023 \\
38	147.616989135742 \\
39	66.5455474853516 \\
40	110.209129333496 \\
41	40.4639930725098 \\
42	107.597686767578 \\
43	115.230834960938 \\
44	138.802337646484 \\
45	47.58984375 \\
46	107.200485229492 \\
47	42.0821189880371 \\
48	101.486320495605 \\
49	135.714248657227 \\
50	110.770439147949 \\
51	171.549728393555 \\
52	75.5082473754883 \\
53	137.144943237305 \\
54	90.558952331543 \\
55	116.777946472168 \\
56	157.422515869141 \\
57	89.5135726928711 \\
58	139.502731323242 \\
59	148.213394165039 \\
60	92.1045761108398 \\
61	96.857551574707 \\
62	143.124481201172 \\
63	145.626800537109 \\
64	159.645568847656 \\
65	118.243370056152 \\
66	69.3945922851562 \\
67	22.9671268463135 \\
68	145.674850463867 \\
69	82.3607711791992 \\
70	61.5435104370117 \\
71	113.096687316895 \\
72	101.142150878906 \\
73	153.61376953125 \\
74	70.590217590332 \\
75	95.2957000732422 \\
76	111.614036560059 \\
77	83.413444519043 \\
78	162.10237121582 \\
79	78.4824523925781 \\
80	71.3355102539062 \\
81	84.4128646850586 \\
82	111.843101501465 \\
83	58.6454620361328 \\
84	95.2463760375977 \\
85	60.0690078735352 \\
86	59.7492866516113 \\
87	87.0835571289062 \\
88	136.919326782227 \\
89	38.32763671875 \\
90	129.335433959961 \\
91	71.7896499633789 \\
92	108.511688232422 \\
93	117.67741394043 \\
94	90.9586944580078 \\
95	110.895820617676 \\
96	33.1515350341797 \\
97	69.4389190673828 \\
98	101.900985717773 \\
99	83.380744934082 \\
};
\addplot [line width=0.5599999999999999pt, black]
table [row sep=\\]{%
0	228.893768310547 \\
1	266.409454345703 \\
2	345.332672119141 \\
3	423.147705078125 \\
4	424.612548828125 \\
5	449.425964355469 \\
6	489.230987548828 \\
7	531.915710449219 \\
8	513.895568847656 \\
9	527.770263671875 \\
10	547.306823730469 \\
11	761.268920898438 \\
12	623.749816894531 \\
13	764.088317871094 \\
14	679.506958007812 \\
15	691.864562988281 \\
16	698.522094726562 \\
17	765.2197265625 \\
18	739.019165039062 \\
19	800.568298339844 \\
20	867.366027832031 \\
21	865.395690917969 \\
22	718.952514648438 \\
23	782.548400878906 \\
24	906.39013671875 \\
25	861.298400878906 \\
26	859.143127441406 \\
27	824.179382324219 \\
28	849.6962890625 \\
29	882.0419921875 \\
30	862.484924316406 \\
31	814.819213867188 \\
32	923.77734375 \\
33	870.294860839844 \\
34	903.366271972656 \\
35	937.39892578125 \\
36	888.8359375 \\
37	883.427978515625 \\
38	947.284790039062 \\
39	868.062927246094 \\
40	917.274597167969 \\
41	859.795776367188 \\
42	928.896118164062 \\
43	941.369934082031 \\
44	971.733825683594 \\
45	883.0322265625 \\
46	966.154357910156 \\
47	901.909790039062 \\
48	961.441040039062 \\
49	995.99462890625 \\
50	971.090942382812 \\
51	1044.26184082031 \\
52	948.598510742188 \\
53	1013.64636230469 \\
54	973.248901367188 \\
55	1000.52899169922 \\
56	1070.65344238281 \\
57	1009.72625732422 \\
58	1064.73461914062 \\
59	1075.72985839844 \\
60	1021.97973632812 \\
61	1038.76232910156 \\
62	1094.41076660156 \\
63	1104.16076660156 \\
64	1145.59924316406 \\
65	1106.32580566406 \\
66	1058.24572753906 \\
67	1012.46752929688 \\
68	1154.10888671875 \\
69	1092.23217773438 \\
70	1076.06213378906 \\
71	1211.93505859375 \\
72	1206.55737304688 \\
73	1262.92358398438 \\
74	1183.85119628906 \\
75	1213.26513671875 \\
76	1236.650390625 \\
77	1253.17431640625 \\
78	1337.54553222656 \\
79	1269.77465820312 \\
80	1267.53137207031 \\
81	1294.75598144531 \\
82	1323.767578125 \\
83	1278.70690917969 \\
84	1321.91625976562 \\
85	1309.08178710938 \\
86	1308.77844238281 \\
87	1341.41015625 \\
88	1412.189453125 \\
89	1318.23083496094 \\
90	1429.97692871094 \\    
91	1482.06311035156 \\
92	1560.01220703125 \\
93	1585.06726074219 \\
94	1629.1474609375 \\
95	1692.21960449219 \\
96	1618.17016601562 \\
97	1705.25512695312 \\
98	1848.88940429688 \\
99	1871.50622558594 \\
};
\end{axis}
\end{tikzpicture}
        \caption{Gradient norm for member and non-member data across all classes}
        \label{fig:grad_classes}
    \end{subfigure}\\[5pt]

    \begin{subfigure}[t]{0.5\textwidth}
        \centering
        \input{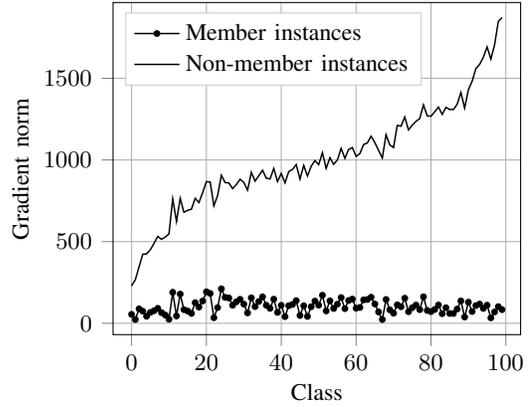}
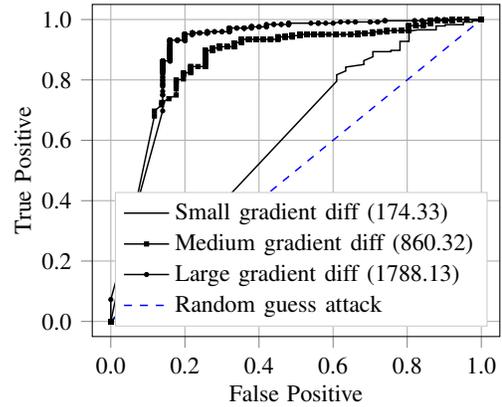
        \caption{Attacker accuracy for class members with various differences in their member/non-member gradient norms}
        \label{fig:attack_classes}
    \end{subfigure}
    \caption{\small Attack accuracy is different for different  output classes (pre-trained CIFAR100-Alextnet model in the stand-alone scenario).}
\end{figure}

Also, we show that the accuracy of our inference attack is higher for classification output classes (of the target model) with a larger difference in member/non-member gradient norms. 
Figure~\ref{fig:grad_classes} plots the average of last layer's gradient norms for different output classes for member and non-member instances; we see that the difference of gradient norms between members and non-members varies across different classes. 
Figure~\ref{fig:attack_classes} shows the receiver operating characteristic (ROC) curve of the inference attack for three output classes with small, medium, and large differences of gradient norm between members and non-members (averaged over many samples). 
As can be seen,  \emph{the larger the difference of gradient norm between members and non-members, the higher the accuracy of the membership inference attack}.

\begin{table}[t!]
    \centering
    \caption{\small Attack accuracy using the outputs of individual activation layers. 
    Pre-trained Alexnet on CIFAR100, stand-alone setting.}
    \label{tab:cifar_diff_layer}
    \begin{tabular}{|c|c|}
        \hline
        \textbf{Output Layer} & \textbf{Attack Accuracy}  \\
        \hline
        Last layer (prediction vector) & $74.6\%$ (black-box)\\
        Second to last &  $74.1\%$\\
        Third to last & $72.9\%$\\
        \hline
        Last three layers, combined & $74.6\%$ \\
        \hline
    \end{tabular}

\end{table}

\begin{table}[t!]
    \centering
    \caption{\small Attack accuracy when we apply the  attack using parameter gradients of different layers. (CIFAR100 dataset with Alexnet architecture, stand-alone scenario)}
    \label{tab:cifar_diff_grads}
    \begin{tabular}{|c|c|}
        \hline
        \textbf{Gradient w.r.t.}  & \textbf{Attack Accuracy}  \\
        \hline
        Last layer parameters & $75.1\%$ \\
        Second to last layer parameters & $74.6\%$\\
        Third to last layer parameters & $73.5\%$\\
        Forth to last layer parameters & $72.6\%$\\
        \hline
        Parameters of last four layers, combined & $75.15\%$ \\
        \hline
    \end{tabular}

\end{table}

\begin{table}[t!]
    \centering
    \caption{\small Attack accuracy using different combinations of layer gradients and outputs. (CIFAR100 dataset with Alexnet architecture, stand-alone scenario)}
    \label{tab:cifar_dif_out_grad}
    \begin{tabular}{|c|c|c|}
        \hline
        \textbf{Gradients w.r.t.} & \textbf{Output Layers}  &  \textbf{Attack Test Accuracy}  \\
        \hline
        Last Layer &  - &$75.10\%$ \\
        Last layer & Last layer & $75.11\%$ \\
        Last Layer & All Layer &  $75.12\%$\\ 
        All  Layer & All Layer &  $75.18\%$\\ 
        \hline
    \end{tabular}
\end{table}

\begin{table}[t!]
\centering
    \caption{\small Attack accuracy for various sizes of the attacker's training dataset. The  size of the target model's training dataset is 50,000. (The CIFAR100 dataset with Alexnet, stand-alone scenario) }
    \label{tab:diff_data_size}
    \begin{tabular}{|c|c|c|}
        \hline
        \textbf{Members Sizes} & \textbf{Non-members Sizes}  & \textbf{Attack Accuracy} \\
        \hline
        10,000 & 2,000 & $73.2\%$ \\
        15,000 & 2,000 & $73.7\%$ \\
        15,000 & 5,000 & $74.8\%$ \\
        25,000 & 5,000 & $75.1\%$\\
        \hline
    \end{tabular}
\end{table}

\begin{table}[t!]
\centering
 \caption{\small Accuracy  of our unsupervised attack compared to the Shadow models approach~\cite{shokri2017membership} for the white-box scenario.}
    \label{tab:unsupervised}
\resizebox{\columnwidth}{!}{%
 \begin{tabular}{|c|c|c|c|}
    \hline
        \multirow{2}{*}{\textbf{Dataset}} &  \multirow{2}{*}{\textbf{Arch}} & \textbf{(Unsupervised) } & \textbf{(Shadow Models)}\\
        &&\textbf{Attack Accuracy}&\textbf{Attack Accuracy}\\
    \hline
        CIFAR100 & Alexnet & $75.0\%$& $70.5\%$ \\
        CIFAR100 & DenseNet & $71.2\%$ & $64.2\%$ \\
        CIFAR100 & ResNet & $63.1\%$ & $60.9\%$ \\
        Texas100 & Fully Connected &$66.3\%$ & $65.3\%$ \\
        Purchase100 & Fully Connected & $71.0\%$ & $68.2\%$ \\
    \hline
        \end{tabular}
}
   
\end{table}

\begin{figure}
        \centering
        \input{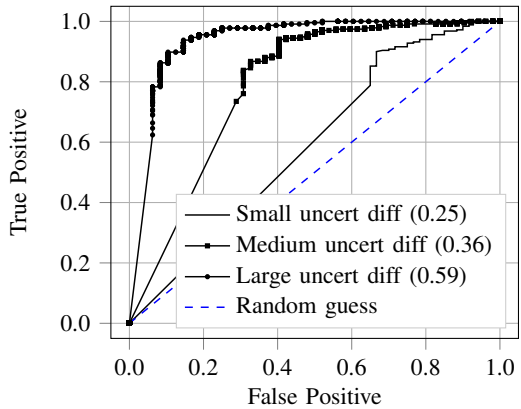}
        \caption{\small Attack's ROC for three different classes of data with large, medium, and small prediction uncertainty values (pre-trained CIFAR100-Alextnet model in the stand-alone scenario).}
        \label{fig:uncertrainty}
\end{figure}

\paragraphbe{Impact of prediction uncertainty:} Previous work~\cite{shokri2017membership} claims that the prediction vector's uncertainty is an important factor in privacy leakage. We validate this claim by evaluating the attack for different classes in CIFAR100-Alexnet with different prediction uncertainties. 
Specifically, we selected three classes with small, medium, and high differences of prediction  uncertainties, where the attack accuracies  are shown in Figure~\ref{fig:uncertrainty} for these classes. 
Similar to the differences in gradient norms, \emph{the classes with higher prediction uncertainty values leak more membership information}.

\begin{table*}[t!]
\centering
    \caption{\small The attack accuracy for different datasets and different target architectures using layer outputs versus gradients.  This is the result of analyzing the stand-alone scenario, where the CIFAR100 models are all obtained from pre-trained online repositories. 
    }
    \begin{tabular}{|c|c|c|c||c|c|c|}
        \hline
        \multicolumn{4} {|c||} {\textbf{Target Model}} &        \multicolumn{3} {c|} {\textbf{Attack Accuracy}}\\
        \hline
        \textbf{Dataset} & \textbf{Architecture}  & \textbf{Train Accuracy} & \textbf{Test Accuracy} & \textbf{Black-box} & \textbf{White-box (Outputs)} & \textbf{White-box (Gradients)} \\
        \hline
        CIFAR100 & Alexnet & $99\%$ & $44\%$ & $74.2\%$ & $74.6\%$ & $75.1\%$ \\
        CIFAR100 & ResNet & $89\%$ & $73\%$ & $62.2\%$  & $62.2\%$ & $64.3\%$\\
        CIFAR100 & DenseNet &$100\%$& $82\%$& $67.7\%$  &$67.7\%$ & $74.3\%$\\
        Texas100 & Fully Connected &$81.6\%$& $52\%$ & $63.0\%$ & $63.3\%$ &$68.3\%$ \\
        Purchase100 & Fully Connected & $100\%$& $80\%$ &$67.6\%$  &$67.6\%$& $73.4\%$ \\
        \hline
    \end{tabular}
    \label{tab:best_white_models}
\end{table*}

\begin{table*}[t!]
    \centering
    \caption{\small Attack accuracy on fine-tuned models. $\set{D}$ is the initial training set, $\set{D}_\Delta$ is the new dataset used for fine-tuning, and $\bar{\set{D}}$ is the set of non-members (which is disjoint with $\set{D}$ and $\set{D}_\Delta$).}
        \label{tab:fine_tune}
     \begin{tabular}{|c|c|c|c|c|c|c|}
        \hline
            \textbf{Dataset} & \textbf{Architecture} & \textbf{Train Acc.} & \textbf{Test Acc.} & \textbf{Distinguish $\set{D}$ from $\set{D}_\Delta$} &  \textbf{Distinguish $\set{D}$ from $\bar{\set{D}}$} & \textbf{Distinguish $\set{D}_\Delta$ from $\bar{\set{D}}$ }\\
        \hline
            CIFAR100 & Alexnet &$100.0\%$ & $39.8\%$& $62.1\%$  & $75.4\%$ & $71.3\%$ \\
            CIFAR100 & DenseNet &$100.0\%$&$64.3\%$ & $63.3\%$ &$74.6\%$ & $71.5\%$\\
            Texas100 & Fully Connected &$95.2\%$&$48.6\%$&$58.4\%$& $68.4\%$ & $67.2\%$\\
            Purchase100 & Fully Connected &$100.0\%$& $77.5\%$& $64.4\%$ &$73.8\%$ & $71.2\%$ \\
        \hline
            \end{tabular}
        
\end{table*}

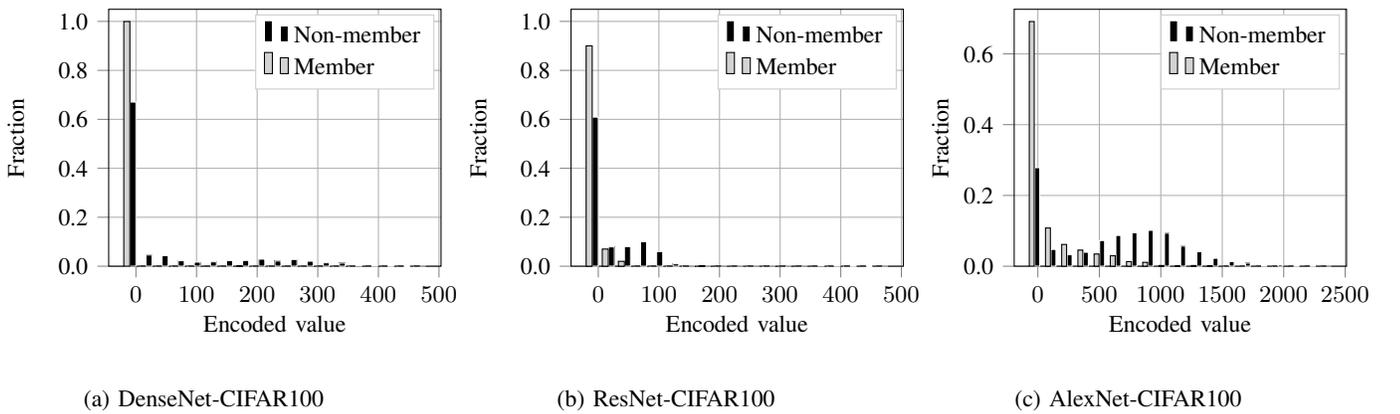
\begin{figure*}[t!]
    \centering
    
    \begin{subfigure}[t]{0.3\textwidth}
        \centering
\begin{tikzpicture}
    \tikzstyle{every node}=[font=\small]

\begin{axis}[
height=\trifigureheight,
legend cell align={left},
legend entries={{Non-member},{Member}},
legend style={draw=white!80.0!black},
tick align=outside,
tick pos=left,
width=\trifigurewidth,
x grid style={lightgray!92.02614379084967!black},
xlabel={Encoded value},
xmajorgrids,
xmin=-44.9210526315789, xmax=503.342105263158,
y grid style={lightgray!92.02614379084967!black},
ylabel={Fraction},
ymajorgrids,
ymin=0, ymax=1.049475,
ytick={0,0.2,0.4,0.6,0.8,1,1.2},
xtick={0,100,200,300,400,500},
yticklabels={0.0,0.2,0.4,0.6,0.8,1.0,1.2}
]
\addlegendimage{ybar,ybar legend,draw=white,fill=black};
\draw[draw=white,fill=black] (axis cs:-10,0) rectangle (axis cs:0,0.6705);
\draw[draw=white,fill=black] (axis cs:16.5789473684211,0) rectangle (axis cs:26.5789473684211,0.046);
\draw[draw=white,fill=black] (axis cs:43.1578947368421,0) rectangle (axis cs:53.1578947368421,0.043);
\draw[draw=white,fill=black] (axis cs:69.7368421052632,0) rectangle (axis cs:79.7368421052632,0.0225);
\draw[draw=white,fill=black] (axis cs:96.3157894736842,0) rectangle (axis cs:106.315789473684,0.017);
\draw[draw=white,fill=black] (axis cs:122.894736842105,0) rectangle (axis cs:132.894736842105,0.018);
\draw[draw=white,fill=black] (axis cs:149.473684210526,0) rectangle (axis cs:159.473684210526,0.023);
\draw[draw=white,fill=black] (axis cs:176.052631578947,0) rectangle (axis cs:186.052631578947,0.0235);
\draw[draw=white,fill=black] (axis cs:202.631578947368,0) rectangle (axis cs:212.631578947368,0.029);
\draw[draw=white,fill=black] (axis cs:229.210526315789,0) rectangle (axis cs:239.210526315789,0.0215);
\draw[draw=white,fill=black] (axis cs:255.789473684211,0) rectangle (axis cs:265.789473684211,0.0275);
\draw[draw=white,fill=black] (axis cs:282.368421052632,0) rectangle (axis cs:292.368421052632,0.0205);
\draw[draw=white,fill=black] (axis cs:308.947368421053,0) rectangle (axis cs:318.947368421053,0.014);
\draw[draw=white,fill=black] (axis cs:335.526315789474,0) rectangle (axis cs:345.526315789474,0.013);
\draw[draw=white,fill=black] (axis cs:362.105263157895,0) rectangle (axis cs:372.105263157895,0.0045);
\draw[draw=white,fill=black] (axis cs:388.684210526316,0) rectangle (axis cs:398.684210526316,0.003);
\draw[draw=white,fill=black] (axis cs:415.263157894737,0) rectangle (axis cs:425.263157894737,0.002);
\draw[draw=white,fill=black] (axis cs:441.842105263158,0) rectangle (axis cs:451.842105263158,0.001);
\draw[draw=white,fill=black] (axis cs:468.421052631579,0) rectangle (axis cs:478.421052631579,0);
\addlegendimage{ybar,ybar legend,draw=black,fill=white!82.74509803921568!black};
\draw[draw=black,fill=white!82.74509803921568!black] (axis cs:-20,0) rectangle (axis cs:-10,0.9995);
\draw[draw=black,fill=white!82.74509803921568!black] (axis cs:6.57894736842105,0) rectangle (axis cs:16.5789473684211,0);
\draw[draw=black,fill=white!82.74509803921568!black] (axis cs:33.1578947368421,0) rectangle (axis cs:43.1578947368421,0);
\draw[draw=black,fill=white!82.74509803921568!black] (axis cs:59.7368421052632,0) rectangle (axis cs:69.7368421052632,0);
\draw[draw=black,fill=white!82.74509803921568!black] (axis cs:86.3157894736842,0) rectangle (axis cs:96.3157894736842,0);
\draw[draw=black,fill=white!82.74509803921568!black] (axis cs:112.894736842105,0) rectangle (axis cs:122.894736842105,0);
\draw[draw=black,fill=white!82.74509803921568!black] (axis cs:139.473684210526,0) rectangle (axis cs:149.473684210526,0);
\draw[draw=black,fill=white!82.74509803921568!black] (axis cs:166.052631578947,0) rectangle (axis cs:176.052631578947,0.0005);
\draw[draw=black,fill=white!82.74509803921568!black] (axis cs:192.631578947368,0) rectangle (axis cs:202.631578947368,0);
\draw[draw=black,fill=white!82.74509803921568!black] (axis cs:219.210526315789,0) rectangle (axis cs:229.210526315789,0);
\draw[draw=black,fill=white!82.74509803921568!black] (axis cs:245.789473684211,0) rectangle (axis cs:255.789473684211,0);
\draw[draw=black,fill=white!82.74509803921568!black] (axis cs:272.368421052632,0) rectangle (axis cs:282.368421052632,0);
\draw[draw=black,fill=white!82.74509803921568!black] (axis cs:298.947368421053,0) rectangle (axis cs:308.947368421053,0);
\draw[draw=black,fill=white!82.74509803921568!black] (axis cs:325.526315789474,0) rectangle (axis cs:335.526315789474,0);
\draw[draw=black,fill=white!82.74509803921568!black] (axis cs:352.105263157895,0) rectangle (axis cs:362.105263157895,0);
\draw[draw=black,fill=white!82.74509803921568!black] (axis cs:378.684210526316,0) rectangle (axis cs:388.684210526316,0);
\draw[draw=black,fill=white!82.74509803921568!black] (axis cs:405.263157894737,0) rectangle (axis cs:415.263157894737,0);
\draw[draw=black,fill=white!82.74509803921568!black] (axis cs:431.842105263158,0) rectangle (axis cs:441.842105263158,0);
\draw[draw=black,fill=white!82.74509803921568!black] (axis cs:458.421052631579,0) rectangle (axis cs:468.421052631579,0);
\end{axis}

\end{tikzpicture}
        \caption{DenseNet-CIFAR100}
    \end{subfigure}\qquad
    \begin{subfigure}[t]{0.3\textwidth}
        \centering
\begin{tikzpicture}
    \tikzstyle{every node}=[font=\small]

\begin{axis}[
height=\trifigureheight,
legend cell align={left},
legend entries={{Non-member},{Member}},
legend style={draw=white!80.0!black},
tick align=outside,
tick pos=left,
width=\trifigurewidth,
x grid style={lightgray!92.02614379084967!black},
xlabel={Encoded value},
xmajorgrids,
xmin=-44.9210526315789, xmax=503.342105263158,
y grid style={lightgray!92.02614379084967!black},
ylabel={Fraction},
ymajorgrids,
ymin=0, ymax=1.049475,
ytick={0,0.2,0.4,0.6,0.8,1,1.2},
xtick={0,100,200,300,400,500},
yticklabels={0.0,0.2,0.4,0.6,0.8,1.0,1.2}
]
\addlegendimage{ybar,ybar legend,draw=white,fill=black};
\draw[draw=white,fill=black] (axis cs:-10,0) rectangle (axis cs:0,0.609);
\draw[draw=white,fill=black] (axis cs:16.5789473684211,0) rectangle (axis cs:26.5789473684211,0.079);
\draw[draw=white,fill=black] (axis cs:43.1578947368421,0) rectangle (axis cs:53.1578947368421,0.08);
\draw[draw=white,fill=black] (axis cs:69.7368421052632,0) rectangle (axis cs:79.7368421052632,0.1);
\draw[draw=white,fill=black] (axis cs:96.3157894736842,0) rectangle (axis cs:106.315789473684,0.06);
\draw[draw=white,fill=black] (axis cs:122.894736842105,0) rectangle (axis cs:132.894736842105,0.01);
\draw[draw=white,fill=black] (axis cs:149.473684210526,0) rectangle (axis cs:159.473684210526,0.0);
\draw[draw=white,fill=black] (axis cs:176.052631578947,0) rectangle (axis cs:186.052631578947,0.0);
\draw[draw=white,fill=black] (axis cs:202.631578947368,0) rectangle (axis cs:212.631578947368,0.0);
\draw[draw=white,fill=black] (axis cs:229.210526315789,0) rectangle (axis cs:239.210526315789,0.0);
\draw[draw=white,fill=black] (axis cs:255.789473684211,0) rectangle (axis cs:265.789473684211,0.0);
\draw[draw=white,fill=black] (axis cs:282.368421052632,0) rectangle (axis cs:292.368421052632,0.0);
\draw[draw=white,fill=black] (axis cs:308.947368421053,0) rectangle (axis cs:318.947368421053,0.0);
\draw[draw=white,fill=black] (axis cs:335.526315789474,0) rectangle (axis cs:345.526315789474,0.0);
\draw[draw=white,fill=black] (axis cs:362.105263157895,0) rectangle (axis cs:372.105263157895,0.0045);
\draw[draw=white,fill=black] (axis cs:388.684210526316,0) rectangle (axis cs:398.684210526316,0.003);
\draw[draw=white,fill=black] (axis cs:415.263157894737,0) rectangle (axis cs:425.263157894737,0.002);
\draw[draw=white,fill=black] (axis cs:441.842105263158,0) rectangle (axis cs:451.842105263158,0.001);
\draw[draw=white,fill=black] (axis cs:468.421052631579,0) rectangle (axis cs:478.421052631579,0);
\addlegendimage{ybar,ybar legend,draw=black,fill=white!82.74509803921568!black};
\draw[draw=black,fill=white!82.74509803921568!black] (axis cs:-20,0) rectangle (axis cs:-10,0.9);
\draw[draw=black,fill=white!82.74509803921568!black] (axis cs:6.57894736842105,0) rectangle (axis cs:16.5789473684211,0.07);
\draw[draw=black,fill=white!82.74509803921568!black] (axis cs:33.1578947368421,0) rectangle (axis cs:43.1578947368421,0.02);
\draw[draw=black,fill=white!82.74509803921568!black] (axis cs:59.7368421052632,0) rectangle (axis cs:69.7368421052632,0);
\draw[draw=black,fill=white!82.74509803921568!black] (axis cs:86.3157894736842,0) rectangle (axis cs:96.3157894736842,0);
\draw[draw=black,fill=white!82.74509803921568!black] (axis cs:112.894736842105,0) rectangle (axis cs:122.894736842105,0);
\draw[draw=black,fill=white!82.74509803921568!black] (axis cs:139.473684210526,0) rectangle (axis cs:149.473684210526,0);
\draw[draw=black,fill=white!82.74509803921568!black] (axis cs:166.052631578947,0) rectangle (axis cs:176.052631578947,0.0005);
\draw[draw=black,fill=white!82.74509803921568!black] (axis cs:192.631578947368,0) rectangle (axis cs:202.631578947368,0);
\draw[draw=black,fill=white!82.74509803921568!black] (axis cs:219.210526315789,0) rectangle (axis cs:229.210526315789,0);
\draw[draw=black,fill=white!82.74509803921568!black] (axis cs:245.789473684211,0) rectangle (axis cs:255.789473684211,0);
\draw[draw=black,fill=white!82.74509803921568!black] (axis cs:272.368421052632,0) rectangle (axis cs:282.368421052632,0);
\draw[draw=black,fill=white!82.74509803921568!black] (axis cs:298.947368421053,0) rectangle (axis cs:308.947368421053,0);
\draw[draw=black,fill=white!82.74509803921568!black] (axis cs:325.526315789474,0) rectangle (axis cs:335.526315789474,0);
\draw[draw=black,fill=white!82.74509803921568!black] (axis cs:352.105263157895,0) rectangle (axis cs:362.105263157895,0);
\draw[draw=black,fill=white!82.74509803921568!black] (axis cs:378.684210526316,0) rectangle (axis cs:388.684210526316,0);
\draw[draw=black,fill=white!82.74509803921568!black] (axis cs:405.263157894737,0) rectangle (axis cs:415.263157894737,0);
\draw[draw=black,fill=white!82.74509803921568!black] (axis cs:431.842105263158,0) rectangle (axis cs:441.842105263158,0);
\draw[draw=black,fill=white!82.74509803921568!black] (axis cs:458.421052631579,0) rectangle (axis cs:468.421052631579,0);
\end{axis}

\end{tikzpicture}
        \caption{ResNet-CIFAR100}
    \end{subfigure}\qquad
 \begin{subfigure}[t]{0.3\textwidth}
        \centering
\begin{tikzpicture}
\tikzstyle{every node}=[font=\small]

\begin{axis}[
height=\trifigureheight,
legend cell align={left},
legend entries={{Non-member},{Member}},
legend style={draw=white!80.0!black},
tick align=outside,
tick pos=left,
width=\trifigurewidth,
y label style={at={(axis description cs:0.11,.5)},anchor=south},
x grid style={lightgray!92.02614379084967!black},
xlabel={Encoded value},
xmajorgrids,
xmin=-192.907894736842, xmax=2511.06578947368,
y grid style={lightgray!92.02614379084967!black},
ylabel={Fraction},
ymajorgrids,
ymin=0, ymax=0.726075,
xtick={0,500,1000,1500,2000,2500},
ytick={0,0.2,0.4,0.6,0.8},
yticklabels={0.0,0.2,0.4,0.6,0.8}
]
\addlegendimage{ybar,ybar legend,draw=white,fill=black};
\draw[draw=white,fill=black] (axis cs:-25,0) rectangle (axis cs:15,0.2785);
\draw[draw=white,fill=black] (axis cs:106.842105263158,0) rectangle (axis cs:146.842105263158,0.0475);
\draw[draw=white,fill=black] (axis cs:238.684210526316,0) rectangle (axis cs:278.684210526316,0.0325);
\draw[draw=white,fill=black] (axis cs:370.526315789474,0) rectangle (axis cs:410.526315789474,0.0395);
\draw[draw=white,fill=black] (axis cs:502.368421052632,0) rectangle (axis cs:542.368421052632,0.0725);
\draw[draw=white,fill=black] (axis cs:634.210526315789,0) rectangle (axis cs:674.210526315789,0.087);
\draw[draw=white,fill=black] (axis cs:766.052631578947,0) rectangle (axis cs:806.052631578947,0.0945);
\draw[draw=white,fill=black] (axis cs:897.894736842105,0) rectangle (axis cs:937.894736842105,0.1015);
\draw[draw=white,fill=black] (axis cs:1029.73684210526,0) rectangle (axis cs:1069.73684210526,0.0935);
\draw[draw=white,fill=black] (axis cs:1161.57894736842,0) rectangle (axis cs:1201.57894736842,0.057);
\draw[draw=white,fill=black] (axis cs:1293.42105263158,0) rectangle (axis cs:1333.42105263158,0.041);
\draw[draw=white,fill=black] (axis cs:1425.26315789474,0) rectangle (axis cs:1465.26315789474,0.0225);
\draw[draw=white,fill=black] (axis cs:1557.10526315789,0) rectangle (axis cs:1597.10526315789,0.013);
\draw[draw=white,fill=black] (axis cs:1688.94736842105,0) rectangle (axis cs:1728.94736842105,0.009);
\draw[draw=white,fill=black] (axis cs:1820.78947368421,0) rectangle (axis cs:1860.78947368421,0.0025);
\draw[draw=white,fill=black] (axis cs:1952.63157894737,0) rectangle (axis cs:1992.63157894737,0.0045);
\draw[draw=white,fill=black] (axis cs:2084.47368421053,0) rectangle (axis cs:2124.47368421053,0.0015);
\draw[draw=white,fill=black] (axis cs:2216.31578947368,0) rectangle (axis cs:2256.31578947368,0.0015);
\draw[draw=white,fill=black] (axis cs:2348.15789473684,0) rectangle (axis cs:2388.15789473684,0);
\addlegendimage{ybar,ybar legend,draw=black,fill=white!82.74509803921568!black};
\draw[draw=black,fill=white!82.74509803921568!black] (axis cs:-70,0) rectangle (axis cs:-30,0.6915);
\draw[draw=black,fill=white!82.74509803921568!black] (axis cs:61.8421052631579,0) rectangle (axis cs:101.842105263158,0.108);
\draw[draw=black,fill=white!82.74509803921568!black] (axis cs:193.684210526316,0) rectangle (axis cs:233.684210526316,0.0615);
\draw[draw=black,fill=white!82.74509803921568!black] (axis cs:325.526315789474,0) rectangle (axis cs:365.526315789474,0.0455);
\draw[draw=black,fill=white!82.74509803921568!black] (axis cs:457.368421052632,0) rectangle (axis cs:497.368421052632,0.0345);
\draw[draw=black,fill=white!82.74509803921568!black] (axis cs:589.210526315789,0) rectangle (axis cs:629.210526315789,0.0295);
\draw[draw=black,fill=white!82.74509803921568!black] (axis cs:721.052631578947,0) rectangle (axis cs:761.052631578947,0.013);
\draw[draw=black,fill=white!82.74509803921568!black] (axis cs:852.894736842105,0) rectangle (axis cs:892.894736842105,0.011);
\draw[draw=black,fill=white!82.74509803921568!black] (axis cs:984.736842105263,0) rectangle (axis cs:1024.73684210526,0.002);
\draw[draw=black,fill=white!82.74509803921568!black] (axis cs:1116.57894736842,0) rectangle (axis cs:1156.57894736842,0.0005);
\draw[draw=black,fill=white!82.74509803921568!black] (axis cs:1248.42105263158,0) rectangle (axis cs:1288.42105263158,0.002);
\draw[draw=black,fill=white!82.74509803921568!black] (axis cs:1380.26315789474,0) rectangle (axis cs:1420.26315789474,0.001);
\draw[draw=black,fill=white!82.74509803921568!black] (axis cs:1512.10526315789,0) rectangle (axis cs:1552.10526315789,0);
\draw[draw=black,fill=white!82.74509803921568!black] (axis cs:1643.94736842105,0) rectangle (axis cs:1683.94736842105,0);
\draw[draw=black,fill=white!82.74509803921568!black] (axis cs:1775.78947368421,0) rectangle (axis cs:1815.78947368421,0);
\draw[draw=black,fill=white!82.74509803921568!black] (axis cs:1907.63157894737,0) rectangle (axis cs:1947.63157894737,0);
\draw[draw=black,fill=white!82.74509803921568!black] (axis cs:2039.47368421053,0) rectangle (axis cs:2079.47368421053,0);
\draw[draw=black,fill=white!82.74509803921568!black] (axis cs:2171.31578947368,0) rectangle (axis cs:2211.31578947368,0);
\draw[draw=black,fill=white!82.74509803921568!black] (axis cs:2303.15789473684,0) rectangle (axis cs:2343.15789473684,0);
\end{axis}

\end{tikzpicture}
        \caption{AlexNet-CIFAR100}
    \end{subfigure}

    \caption{The distribution of the encoded values (i.e., the attack output) for the member and non-member instances of our unsupervised algorithm are distinguishable. This is the intuition behind the high accuracy of our unsupervised attack.}
    \label{fig:clustering}
\end{figure*}

\subsection{Stand-Alone Setting: Unsupervised Attacks} 
We also implement our attacks in an unsupervised scenario, in which 
the attacker has data points sampled from the same underlying distribution, but he does not know their  member and non-member labels. 
In this case, the attacker classifies the tested records into two clusters as described in Section~\ref{sec:inference_supervised_unsupervised}.

We implemented our attack and compared its performance to Shadow models of 
Shokri et al.~\cite{shokri2017membership} introduced earlier. 
We train our unsupervised models on various datasets based on the training and test dataset sizes in Table~\ref{tab:datasize}.
We train a single Shadow model on each of Texas100 and Purchase100 datasets  using training sizes according to  Table~\ref{tab:datasize}. The training sets of the Shadow models do no overlap with the training sets of the target models. 
For the CIFAR100 dataset, however, our Shadow model uses a training dataset that overlaps with the target model's dataset, as we do not have enough  instances (we train each model with 50,000 instances out of the total  60,000 available records).

After the training,  we use the Spectral clustering algorithm~\cite{von2007tutorial} to divide the input samples into two clusters.
As shown earlier (Figure~\ref{fig:grad_norms}), the member instances have smaller gradient norm values.
Therefore, we assign the member label to the cluster with a smaller average gradient norm, and the non-member label to the other cluster.  

Table~\ref{tab:unsupervised} compares the accuracy of our unsupervised attack with shadow training~\cite{shokri2017membership} on various datasets and architectures. We see that \emph{our approach offers a noticeably higher accuracy}. 

The intuition behind our attack working is that the encoded values of our unsupervised algorithm present  different distributions for member and non-member samples. This can be seen in Figure~\ref{fig:clustering} for various datasets and architectures.

\subsection{Stand-Alone Setting: Attacking Fine-Tuned Models}

We investigate privacy leakage of fine-tuned target models. 
In this scenario, the victim trains a model with dataset $D$, then he uses a dataset $D_\Delta$ to fine-tune the trained model to improve its performance. Hence, the attacker has two snapshots of the trained model, one using only $D$, and one for the same model which is fine-tuned using  $D_\Delta$.
We assume the attacker has access to both of the trained models (before and after fine-tuning). 
We are interested in applying the membership inference attack in this scenario, where the goal of the adversary is to distinguish between the members of $D$, $D_\Delta$, and $\bar{D}$, which is a set of non-members.  

We use the same training dataset as in the previous experiments (Table~\ref{tab:datasize}); we used $60\%$ of the train dataset as $D$ and the rest for $D_\Delta$.
Table~\ref{tab:fine_tune} shows the train, test, and attack accuracy for different scenarios.
As can be seen, the attacker is able to distinguish between members (in  $D$ or $D_\Delta$) and non-members ($\bar{D}$) with  accuracies similar to previous settings. 
Additionally, the attacker can also distinguish between the members of $D$ and  $D_\Delta$ with reasonably high accuracies.

\subsection{Federated Learning Settings: Passive Inference Attacks}

Table~\ref{tab:datasize_fed} shows the dataset sizes used in our  federated attack experiments. 
For the CIFAR100 experiment with a local attacker, each participant uses  30,000 instances to train, which overlaps between various participants due to non-sufficient number of instances. For all the other experiments, the participants use non-overlapping datasets. In the following, we present the attack in various settings.

\begin{table*}
    \centering
    \caption{\small Attack accuracy in the federated  learning setting. There are 4 participants. A global attacker is the central parameter aggregator, and the local attacker is one of the participants. The global attacker performs the inference against each individual participant, and we report the average attack accuracy. The local attacker performs the inference against all other participants. The passive attacker follows the protocol and only observes the updates. The active attacker changes its updates, or (in the case of a global attack) isolates one participant by not passing the updates of other participants to it, in order to increase the information leakage. }
    \label{tab:fed_all}
\begin{tabular}{|c|c||c|c|c|c||c|c|}
    \hline
    \multicolumn{2}{ |c| |}{\textbf{Target Model}} &  \multicolumn{4}{ c|| }{\textbf{Global Attacker (the parameter aggregator)}}&  \multicolumn{2}{ c| }{\textbf{Local Attacker (a participant)}}  \\ 
    \cline{3-8}
    \multicolumn{2}{| c|| }{}&Passive& \multicolumn{3}{ c|| }{Active}&Passive & Active \\
    \cline{1-2}\cline{4-6}\cline{8-8}
    Dataset & Architecture &  & Gradient Ascent  & Isolating &  Isolating Gradient Ascent &  & Gradient Ascent \\
    \hline
    CIFAR100 & Alexnet & $85.1\%$ & $88.2\%$ & $89.0\%$ &$92.1\%$  & $73.1\%$ & $76.3\%$ \\
    CIFAR100 & DenseNet& $79.2\%$ & $82.1\%$ & $84.3\%$ & $87.3\%$& $72.2\%$& $76.7\%$\\
    Texas100 & Fully Connected& $66.4\%$ & $69.5\%$ & $69.3\%$&$71.7\%$& $62.4\%$& $66.4\%$\\
    Purchase100 & Fully Connected & $72.4\%$& $75.4\%$ &$75.3\%$& $82.5\%$ & $65.8\%$& $69.8\%$\\
    \hline
\end{tabular}
\end{table*}

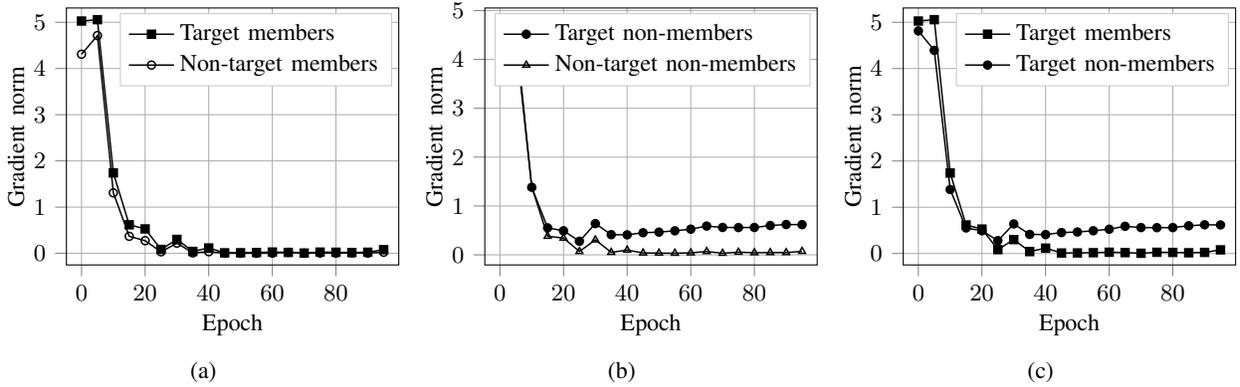
\begin{figure*}[t!]
    \centering
    \begin{subfigure}[t]{0.3\textwidth}
        \centering
\begin{tikzpicture}
    \tikzstyle{every node}=[font=\small]

\begin{axis}[
height=\trifigureheight,
legend cell align={left},
y label style={at={(axis description cs:0.18,.5)},anchor=south},
legend entries={{Target members},{Non-target members}},
legend style={draw=white!80.0!black},
tick align=outside,
tick pos=left,
width=\trifigurewidth,
x grid style={lightgray!92.02614379084967!black},
xlabel={Epoch},
xmajorgrids,
xmin=-4.75, xmax=99.75,
y grid style={lightgray!92.02614379084967!black},
ylabel={Gradient norm},
ymajorgrids,
ytick={0,1,...,5},
ymin=-0.251769961090758, ymax=5.30923075624742
]
\addplot [line width=0.5599999999999999pt, black, mark=square*, mark size=1.5, mark options={solid}]
table [row sep=\\]{%
0	5.02667188644409 \\
5	5.05645799636841 \\
10	1.74039196968079 \\
15	0.616923570632935 \\
20	0.528172791004181 \\
25	0.0839772075414658 \\
30	0.301171332597733 \\
35	0.0409685336053371 \\
40	0.116320580244064 \\
45	0.0102091608569026 \\
50	0.0119711011648178 \\
55	0.0172107126563787 \\
60	0.0274020228534937 \\
65	0.0191987082362175 \\
70	0.00501802284270525 \\
75	0.0260684341192245 \\
80	0.022965706884861 \\
85	0.0158822443336248 \\
90	0.0232855323702097 \\
95	0.0788435488939285 \\
};
\addplot [line width=0.5599999999999999pt, black, mark=o, mark size=1.5, mark options={solid}]
table [row sep=\\]{%
0	4.30851697921753 \\
5	4.712965965271 \\
10	1.30870354175568 \\
15	0.368230223655701 \\
20	0.272874623537064 \\
25	0.0322835445404053 \\
30	0.22018951177597 \\
35	0.00906867813318968 \\
40	0.0345387533307076 \\
45	0.00245577096939087 \\
50	0.00205908017233014 \\
55	0.00100279878824949 \\
60	0.00274417852051556 \\
65	0.020432323217392 \\
70	0.00213117245584726 \\
75	0.00596591643989086 \\
80	0.00275918724946678 \\
85	0.00799521990120411 \\
90	0.00470739835873246 \\
95	0.022746454924345 \\
};
\end{axis}

\end{tikzpicture}
        \caption{}
        \label{fig:train_target_train_nontarget}
    \end{subfigure}
        \begin{subfigure}[t]{0.3\textwidth}
        \centering
\begin{tikzpicture}
    \tikzstyle{every node}=[font=\small]

\begin{axis}[
height=\trifigureheight,
legend cell align={left},
legend entries={{Target non-members},{Non-target non-members}},
legend style={draw=white!80.0!black},
tick align=outside,
tick pos=left,
width=\trifigurewidth,
x grid style={lightgray!92.02614379084967!black},
y label style={at={(axis description cs:0.18,.5)},anchor=south},
xlabel={Epoch},
xmajorgrids,
xmin=-4.75, xmax=99.75,
y grid style={lightgray!92.02614379084967!black},
ylabel={Gradient norm},
ymajorgrids,
ytick={0,1,...,5},
ymin=-0.207055754587054, ymax=5.05254353545606
]
\addplot [line width=0.5599999999999999pt, black, mark=*, mark size=1.5, mark options={solid}]
table [row sep=\\]{%
0	4.8134708404541 \\
5	4.39297199249268 \\
10	1.38379395008087 \\
15	0.552519619464874 \\
20	0.491988688707352 \\
25	0.278049141168594 \\
30	0.639399349689484 \\
35	0.412834078073502 \\
40	0.409231960773468 \\
45	0.452589094638824 \\
50	0.464814871549606 \\
55	0.491826683282852 \\
60	0.525444269180298 \\
65	0.588499903678894 \\
70	0.56047385931015 \\
75	0.55843311548233 \\
80	0.560200810432434 \\
85	0.59951788187027 \\
90	0.620666980743408 \\
95	0.617851972579956 \\
};
\addplot [line width=0.5599999999999999pt, black, mark=triangle, mark size=1.5, mark options={solid}]
table [row sep=\\]{%
0	4.40911865234375 \\
5	4.56990671157837 \\
10	1.38853061199188 \\
15	0.380504190921783 \\
20	0.342713356018066 \\
25	0.0689850002527237 \\
30	0.305819392204285 \\
35	0.0504038780927658 \\
40	0.0974165797233582 \\
45	0.0383830182254314 \\
50	0.0354956798255444 \\
55	0.0324388891458511 \\
60	0.0407826788723469 \\
65	0.0680200159549713 \\
70	0.0320169404149055 \\
75	0.0540318377315998 \\
80	0.0431632921099663 \\
85	0.0462457127869129 \\
90	0.046037320047617 \\
95	0.0711009129881859 \\
};
\end{axis}
\end{tikzpicture}
        \caption{}
        \label{fig:test_target_test_nontarget}
    \end{subfigure}
    \begin{subfigure}[t]{0.3\textwidth}
        \centering
\begin{tikzpicture}
    \tikzstyle{every node}=[font=\small]

\begin{axis}[
height=\trifigureheight,
legend cell align={left},
legend entries={{Target members},{Target non-members}},
legend style={draw=white!80.0!black},
tick align=outside,
tick pos=left,
width=\trifigurewidth,
y label style={at={(axis description cs:0.18,.5)},anchor=south},
x grid style={lightgray!92.02614379084967!black},
xlabel={Epoch},
xmajorgrids,
xmin=-4.75, xmax=99.75,
y grid style={lightgray!92.02614379084967!black},
ylabel={Gradient norm},
ymajorgrids,
ytick={0,1,...,5},
ymin=-0.24755397583358, ymax=5.30902999504469
]

\addplot [line width=0.5599999999999999pt, black, mark=square*, mark size=1.5, mark options={solid}]
table [row sep=\\]{%
0	5.02667188644409 \\
5	5.05645799636841 \\
10	1.74039196968079 \\
15	0.616923570632935 \\
20	0.528172791004181 \\
25	0.0839772075414658 \\
30	0.301171332597733 \\
35	0.0409685336053371 \\
40	0.116320580244064 \\
45	0.0102091608569026 \\
50	0.0119711011648178 \\
55	0.0172107126563787 \\
60	0.0274020228534937 \\
65	0.0191987082362175 \\
70	0.00501802284270525 \\
75	0.0260684341192245 \\
80	0.022965706884861 \\
85	0.0158822443336248 \\
90	0.0232855323702097 \\
95	0.0788435488939285 \\
};
\addplot [line width=0.5599999999999999pt, black, mark=*, mark size=1.5, mark options={solid}]
table [row sep=\\]{%
0	4.8134708404541 \\
5	4.39297199249268 \\
10	1.38379395008087 \\
15	0.552519619464874 \\
20	0.491988688707352 \\
25	0.278049141168594 \\
30	0.639399349689484 \\
35	0.412834078073502 \\
40	0.409231960773468 \\
45	0.452589094638824 \\
50	0.464814871549606 \\
55	0.491826683282852 \\
60	0.525444269180298 \\
65	0.588499903678894 \\
70	0.56047385931015 \\
75	0.55843311548233 \\
80	0.560200810432434 \\
85	0.59951788187027 \\
90	0.620666980743408 \\
95	0.617851972579956 \\
};
\end{axis}
\end{tikzpicture}
        \caption{}
        \label{fig:test_target_train_target}
    \end{subfigure}%

    \caption{The impact of the global active gradient ascent attack on the target model's training process. Figures show the gradient norms of various instances (Purchase100 dataset) during the training phase, while the target instances are under attack.}
    \label{fig:active_res}
\end{figure*}

\begin{table}
    \centering
    \caption{\small Dataset sizes in the federated learning experiments}
    \label{tab:datasize_fed}
    \resizebox{\columnwidth}{!}{%
    \begin{tabular}{|c|c|c|c|c|c|c|}
        \hline
        & \multicolumn{2}{ c| }{\textbf{Parties' Datasets}} &  \multicolumn{4}{ c| }{\textbf{Inference Attack Model}} \\ \cline{2-7}
        \multirow{2}{*}{\textbf{Datasets}} & \multirow{2}{*}{\textbf{Training}}   & \multirow{2}{*}{\textbf{Test}}  & \textbf{Training} & \textbf{Training} &\textbf{Test } & \textbf{Test}  \\
        &&&\textbf{Members} &\textbf{Non-members}&\textbf{Members} &\textbf{Non-members}\\
        \hline
        CIFAR100 & 30,000 & 10,000& 15,000 & 5,000 & 5,000& 5,000\\  
        Texas100 & 8,000 & 70,000 & 4,000& 4,000 & 4,000& 4,000\\
        Puchase100 & 10,000 & 50,000&5,000&5,000&5,000&5,000 \\
        \hline
    \end{tabular}
    }
    
\end{table}

\begin{table}
\centering
\caption{\small The accuracy of the passive global attacker in the federated setting when the attacker uses various training epochs. 
 (CIFAR100-Alexnet)}
    \label{tab:different_epochs}
\begin{tabular}{|c|c|}
    \hline
        \textbf{Observed Epochs} & \textbf{Attack Accuracy} \\
    \hline
$5,10,15,20,25$ & $57.4\%$ \\
$10,20,30,40,50$ & $76.5\%$ \\
$50,100,150,200,250$ & $79.5\%$ \\
$100,150,200,250,300$ & $85.1\%$  \\
    \hline
        \end{tabular}
    
\end{table}

\begin{table}
\centering
\caption{\small The accuracy of the passive local attacker for different numbers of participants. (CIFAR100-Alexnet)}
\label{tab:different_parties}
        \vspace*{-0.1cm}
\begin{tabular}{|c|c|}
    \hline
        \textbf{Number of Participants}  &\textbf{Attack Accuracy} \\
              \hline
    2 & $89.0\%$ \\
    3& $78.1\%$\\
    4 &$76.7\%$ \\
    5 & $67.2\%$ \\
        \hline
        \end{tabular}
        \vspace*{-0.5cm}
\end{table}

\paragraphbe{The Passive Global Attacker:} 
In this scenario, the attacker (the parameter aggregator) has access to the target model's parameters over multiple training epochs (see Section~\ref{sec:inference_standalone_federated}).  Thus, he can passively collect all the parameter updates from all of the participants, at each epoch, and can perform the membership inference attack against each of the target participants, separately.

Due to our limited GPU resources, our attack observes each target participant during only five (non-consecutive) training epochs.
Table~\ref{tab:different_epochs} shows the accuracy of our attack when it uses different sets of training epochs (for the CIFAR100 dataset with Alexnet). 
We see that \emph{using later epochs, substantially increases the attack accuracy}.
Intuitively, this is because the earlier training epochs contain information of the  generic features of the dataset, which do not  leak significant membership information, however, the later epochs contain more membership information as the model starts  to learn the outliers in such epochs~\cite{zhang2016understanding}.

Table~\ref{tab:fed_all} presents the results of this attack on different datasets. 
For the Purchase100 and Texas100 datasets we use the $[40,60,80,90,100]$ training epochs, and for the CIFAR100 dataset we use epochs  $[100,150,200,250,300]$. 
When the attacker has access to several training epochs in the CIFAR100 target models, he achieves a high membership attack accuracy. In Texas100 and Purchase100 datasets, however, the accuracy of the attack decreases compare to the stand-alone setting. This is due to the fact that \emph{averaging in the federated learning scenarios will reduce the impact of each individual party}.

\paragraphbe{The Passive Local Attacker:}  A local attacker cannot observe the model updates of the participants; he can only observe the \emph{aggregate} model parameters. 
We use the same attack model architecture as that of the global attack.
In our experiments, there are four participants (including the local attacker). The goal of the attacker is to learn if a target input has been a member of the training data of any other participants. 
Table~\ref{tab:fed_all} shows the  accuracy of our attack on various datasets. 
As expected, a local  attack has a  lower accuracy compared to the  global attack; this is because the local attacker observes the \emph{aggregate} model parameters of all participants, which limits the extent of membership leakage.  \emph{The accuracy of the local attacker  degrades for larger numbers of participants}.
This is shown in  Table~\ref{tab:different_parties} for the CIFAR100 on Alexnet model.

\subsection{Federated Learning Settings: Active Inference Attacks}

Table~\ref{tab:fed_all} shows the results of attacks on federated learning. 

\paragraphbe{The Gradient Ascent Attacker:} In this scenario, the attacker adversarially manipulates the learning process to improve the membership inference accuracy. The active attack is described in Section~\ref{sec:inference_passive_active}.  We evaluate the attack accuracy on predicting the membership of 100 randomly sampled member instances, from the target model, and  100 non-member instances. 
For all such target instances (whose membership is unknown to the attacker), the attacker updates their data features towards ascending the gradients of the global model (in case of the global attack) or the local model (in the case of a local attack).

Figure~\ref{fig:active_res} compares the last-layer gradient norm of the target model for different data points. As Figure~\ref{fig:train_target_train_nontarget} shows, when the attacker ascends on the gradients of the target instances, the gradient norm of the target members will be very similar to that of non-target member instances in various training epochs. 
On the other hand, this is not true for the non-member instances as shown in Figure~\ref{fig:test_target_test_nontarget}. 

Intuitively, this is because applying the gradient ascent algorithm on a member instance will trigger the target model to try to minimize its loss by descending in the direction of the model's gradient for those instances (and therefore nullify the effect of the attacker's ascent). For target non-member instances, however, the  model will not explicitly change their gradient, as they do not influence the training loss function. The attacker repeats gradient ascend algorithm for each epoch of the training, therefore, the gradient of the  model will keep increasing  on such non-member instances. Figure~\ref{fig:test_target_train_target} depicts the resulted distinction between the gradient norm of the member and non-member target instances. The active gradient ascend attacker forces the target model to behave drastically different between target member and target non-member instances which makes the membership inference attack easier. As a result, compared to the passive global attacker we see that the active attack can noticeably gain higher accuracy. In the local case, the accuracy is lower than the global attack due to the observation of aggregated parameters from multiple participants.

\paragraphbe{The Isolating Attacker:}  The parameter aggregation in the federated learning scenario negatively influences the accuracy of the membership inference attacks. An active global attacker can overcome this problem by isolating a target participant, and creating a local view of the network for it. In this scenario, the attacker does not send the aggregate parameters of all parties to the target party. Instead, the attacker isolates the target participant and segregates the target participant's learning process. 

When the attacker isolates the target participant, then the target participant's model  does not get aggregated with the parameters of other parties. As a result, it stores more information about its training dataset in the model.  Thus, simply \emph{isolating the training of a target model significantly increases the attack accuracy}. We can apply the isolating method to the gradient ascent attacker and further improve the attacker accuracy. See Table~\ref{tab:fed_all} for all the results. 

\section{Related Work}
Investigating different inference attacks on deep neural networks is an active area of research. 

\subsection{Membership Inference Attacks}

Multiple research papers have studied  membership inference attacks  in a black-box setting~\cite{shokri2017membership, long2018understanding, yeom2018privacy}.  Homer et al.~\cite{homer2008resolving} performed one of the first membership inference attacks on genomic data.  Shokri et al.~\cite{shokri2017membership} showed that an ML model's output has distinguishable properties about its training data, which could be exploited by the adversary's inference model. They  introduced shadow models that mimic the behavior of the target model, which are used by the attacker to train the attack model. Salem et al.~\cite{salem2018ml} extended the attacks of Shokri et al.~\cite{shokri2017membership} and  showed  empirically that it is possible to use a single shadow model (instead of several shadow models used in ~\cite{shokri2017membership}) to perform the same attack. They further demonstrated that even if the attacker does not have access to the target model's training data, she can use statistical properties of outputs (e.g., entropy) to perform membership inference.
Yeom et al.~\cite{yeom2018privacy} demonstrated the relationship between overfitting and membership inference attacks. Hayes et al.~\cite{hayes2017logan} used generative adversarial networks to perform membership attacks on generative models.  

Melis et al.~\cite{melis2018inference} developed a new set of membership inference attacks for the collaborative learning. The attack assumes that the participants update the central server after each mini-batch, as opposed to  updating after each training epoch~\cite{shokri2015privacy, mcmahan2016communication}. Also, the proposed membership inference attack is designed exclusively for models that use explicit word embeddings (which reveal the set of words used in the training sentences in a mini-batch) with very small training mini-batches. 

In this paper, we evaluate standard learning mechanisms for deep learning and standard target models for various architectures. We showed that our attacks work even if we use pre-trained, state-of-the-art target models.

Differential privacy~\cite{dwork2006calibrating, dwork2014algorithmic} has been used as a strong defense mechanism against inference attacks in the context of machine learning~\cite{chaudhuri2011differentially, bassily2014private, abadi2016deep, papernot2018scalable}.  Several works~\cite{nasr2018machine, hamm2017minimax, huang2018generative} have shown that by using adversarial training, one can find a better trade-off between privacy and model accuracy. However, the focus of this line of work is on the membership inference attack in the black-box setting. 

\subsection{Other Inference Attacks} 

An attacker with additional information about the training data distribution can perform various types of inference attacks. Input inference~\cite{fredrikson2015model}, attribute inference~\cite{carlini2018secret}, parameter inference~\cite{tramer2016stealing, wang2018stealing}, and side-channel attacks~\cite{wei2018know} are several examples of such attacks. 
Ateniese et al.~\cite{ateniese2015hacking} show that  an adversary with access to the parameters of machine learning models such as Support Vector Machines (SVM) or Hidden Markov Models (HMM)~\cite{robert2014machine}  can extract valuable information about the training data (e.g., the accent of the speakers in speech recognition models).

\section{Conclusions}
We designed and evaluated novel white-box membership inference attacks against neural network models by exploiting the privacy vulnerabilities of the stochastic gradient descent algorithm. 
We demonstrated our attacks in the stand-alone and federated settings, with respect to passive and active inference attackers, and assuming different adversary prior knowledge. 
We showed that even well-generalized models are significantly susceptible to such white-box membership inference attacks.  Our work did not investigate theoretical bounds on the privacy leakage of deep learning in the white-box setting, which would remain as a topic of future research. 

\section*{Acknowledgments}

This work was supported in part by the NSF grant CNS-$1525642$, as well as the Singapore Ministry of Education Academic Research Fund Tier~1, R-252-000-660-133. Reza Shokri would like to acknowledge the support of NVIDIA
Corporation with the donation of a Titan Xp GPU which was used for this
research.


\appendices

\section{Architecture of the Attack Model}\label{sec:appendix}
\begin{table}[h!]
    \centering
    \caption{Attack model layer sizes}
    \label{tab:model_sizes}
    \resizebox{\columnwidth}{!}{%

    \begin{tabular}{|c|c|c|}
        \hline
        \textbf{Name} & \textbf{Layers} & \textbf{Details}\\
        \hline
        \multirow{3}{*}{Output Component} & \multirow{3}{*}{2 Fully Connected Layers} &  Sizes: $128$, $64$  \\
        && Activation: ReLU \\
        && Dropout: 0.2\\
        \hline
        \multirow{3}{*}{Label Component} & \multirow{3}{*}{2 Fully Connected Layers} &  Sizes: $128$, $64$  \\
        && Activation: ReLU \\
         && Dropout: 0.2\\
        \hline
        \multirow{3}{*}{Loss Component} & \multirow{3}{*}{2 Fully Connected Layers} &  Sizes: $128$, $64$  \\
        && Activation: ReLU \\
                && Dropout: 0.2\\
        \hline
        \multirow{6}{*}{Gradient Component} & \multirow{4}{*}{Convolutional Layer}& Kernels: $1000$ \\
        && Kernel size: $1\times $ Next layer \\
        && Stride:1\\
        && Dropout: 0.2\\
        \cline{2-3}
        &\multirow{3}{*}{2 Fully Connected Layers}&Sizes: $128$, $64$  \\
        && Activation: ReLU \\
                && Dropout: 0.2\\
        \hline
        \multirow{3}{*}{Encoder Component} & \multirow{3}{*}{4 Fully Connected Layers} &  Sizes: $256$, $128$, $64$, $1$  \\
        && Activation: ReLU \\
        && Dropout: 0.2\\
        \hline
        \multirow{3}{*}{Decoder Component} & \multirow{3}{*}{2 Fully Connected Layers} &  Sizes: $64$, $4$  \\
        && Activation: ReLU \\
        && Dropout: 0.2\\
        \hline
    \end{tabular}
    }
    
\end{table}

\end{document}